\pdfoutput=1
\documentclass[10pt,twocolumn,letterpaper]{article}
\usepackage[pagenumbers]{cvpr}
\usepackage{graphicx}
\usepackage{amsmath,amssymb,amsthm}
\usepackage{booktabs,multirow,array}
\usepackage{longtable,pdflscape}
\usepackage{xcolor}
\usepackage{tikz}
\usetikzlibrary{arrows.meta,calc,positioning,fit,backgrounds}
\usepackage{microtype}
\usepackage{algorithm}
\usepackage{placeins}
\floatstyle{boxed}\restylefloat{algorithm}

\definecolor{draftred}{RGB}{190,20,30}

\newcommand{\Id}{\mathrm{Id}}
\newcommand{\round}[1]{\operatorname{round}(#1)}
\newcommand{\enc}{E'}
\newcommand{\dec}{D'}
\newif\ifdraftnotes
\draftnotesfalse %

\usepackage{iftex}
\ifXeTeX
  \usepackage{fontspec}
\fi

\newif\ifincludeappendix
\includeappendixtrue %
\definecolor{codecblue}{HTML}{315D80}
\definecolor{codeclight}{HTML}{E6EDF3}
\definecolor{codecgray}{HTML}{EFEFEF}

\definecolor{cvprblue}{rgb}{0.21,0.49,0.74}
\usepackage[pagebackref,breaklinks,colorlinks,allcolors=cvprblue]{hyperref}
\title{Your Codec, Your Generator:\\Idempotent Generative Compression without Retraining}
\author{Raz Haim\qquad\qquad Assaf Shocher\\
Technion -- Israel Institute of Technology
}
\makeatletter
\patchcmd{\@maketitle}{\vskip .375in}{\vskip 0pt}{}{\PackageError{paper}{Title top-space patch failed}{Check the title macro.}}
\patchcmd{\@maketitle}{\vspace*{24pt}}{\vspace*{8pt}}{}{\PackageError{paper}{Title-author space patch failed}{Check the title macro.}}
\g@addto@macro\@maketitle{%
  \begingroup
  \centering
  \vspace{-0.7cm}
  \includegraphics[width=0.96\textwidth]{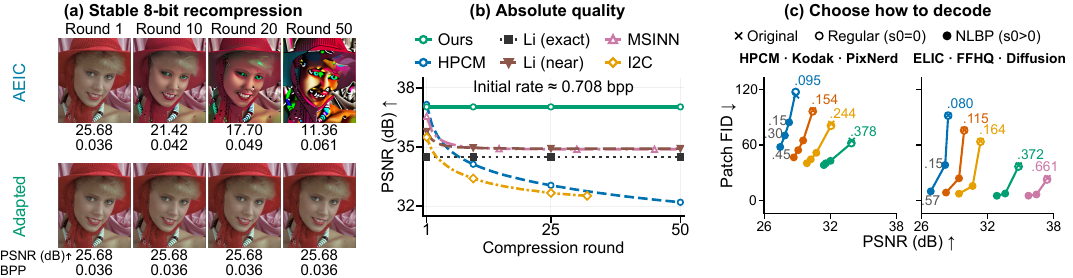}\par
  \par\vspace{-0.1cm}\captionof{figure}{\small
(a) 1, 10, 20, and 50 recompressions, with and without our adapter. (b) PSNR across recompression rounds against methods designed for idempotency, at a common initial bitrate, on Kodak benchmark. (c) Nonlinear back-projection controls the perception--distortion trade-off. Each colored curve varies decoding while keeping the transmitted bitstreams fixed. Top labels indicate bitrate in bpp.
  }
  \label{fig:teaser}
  \endgroup
  \vspace{4pt}
}
\makeatother
\begin{document}
\maketitle

\setlength{\parskip}{\dimexpr\parskip\relax}
\patchcmd{\endabstract}{\vspace*}{\vspace}{}{\PackageError{paper}{Abstract trailing-space patch failed}{Check the abstract macro.}}
\begin{abstract}
We introduce an image compression framework with three properties:
(i)~adaptation of pretrained codecs without retraining or fine-tuning,
(ii)~cycle-consistent recompression, even of already-encoded images, and
(iii)~decoding with independently chosen generative priors.
We wrap the frozen encoder with non-learned operations so that every continuous latent has a preimage, yielding a surjective pseudo-invertible neural network. A nonlinear right inverse built from the pretrained decoder satisfies $E'(D'(E'(x)))=E'(x)$.
To extend consistency to quantized bitstreams, the receiver searches for a continuous latent that reproduces both the transmitted main symbols and entropy side information. A successful search and an exact correction preserve both streams under deterministic coding; the final reconstruction is checked through the actual coding path.
The encoder's pseudo-inverse enables a nonlinear counterpart of back-projection. In the linear case, back-projection has the closed-form solution $x'=x+A^\dagger(y-Ax)$, giving the closest reconstruction to $x$ satisfying $Ax'=y$. Our nonlinear formulation enforces the transmitted code on proposals from an independently chosen generative model. Adjusting the starting noise controls the prior's influence without changing that code.
Experiments show small first-pass quality changes, higher retained quality than trained consistency codecs over 50 recompressions, also under 8-bit storage, and decoder-side control of the perception--distortion trade-off. Decoupling the codec from its prior lets the receiver adopt future generators without retraining or re-encoding.
\par
\end{abstract}

\setlength{\parskip}{\dimexpr\parskip\relax}
\patchcmd{\cvprsection}{-10pt plus -2pt minus -2pt}{-10pt}{}{\PackageError{paper}{Introduction heading patch failed}{Check the section macro.}}
\section{Introduction}
\label{sec:intro}
Image compression underpins the storage and exchange of visual information at global scale. Images are repeatedly decoded and recompressed as they are saved, shared, and archived, so preserving quality requires more than a good first reconstruction. After the initial loss, further compression cycles should leave the image unchanged. This property is \emph{idempotency}: for a reconstruction operator $T$ with fixed codec settings and deterministic decoding, $T(T(x))=T(x)$. Many learned codecs instead accumulate degradation with every round~\cite{instability}. We seek to preserve both the reconstructed image and its complete transmitted code.\par

Code consistency fixes what must be preserved, while leaving unrecorded detail available to a generative prior. Large generative models capture rich knowledge of image structure and natural-image distributions~\cite{ddpm,ldm}. This knowledge could improve decoding, including for images compressed before a stronger model becomes available. We would like to choose such a prior at decoding time, without retraining the codec or changing the bitstream. The challenge is to let the generator supply plausible detail while respecting everything the code specifies.

The linear setting connects these goals. When a linear encoder can produce every latent in its target space, it is surjective, and its pseudo-inverse is a right inverse. This supports exact code recovery and closed-form back-projection: the smallest correction to an image that makes it match given measurements. Surjective Pseudo-Invertible Neural Networks (SPNN)~\cite{spnn} extend this structure to nonlinear maps through a right inverse and a bijective completion.

We adapt a pretrained codec without training: a wrapper turns its frozen encoder into an SPNN and builds a nonlinear right inverse from its frozen decoder~\cite{dejavae}. The continuous transforms satisfy
\vspace{-0.1cm}
\begin{equation}
 E'(D'(E'(x)))=E'(x).
 \label{eq:intro-consistency}
\end{equation}
A continuous right inverse alone does not make a codec idempotent: quantization changes the latent, and re-encoding can change both the main symbols and their entropy side information. Our adaptation searches for a continuous representative of the received symbol pair, writes it into the reconstruction, and verifies the complete coding path. A successful check certifies code preservation under deterministic coding. The first encode remains the original codec's, so the receiver can also adapt already-encoded images without changing their initial bitstream.

Generative decoding then becomes a nonlinear inverse problem with the encoder as the degradation. Prior training-free decoders enforce the code approximately, by guidance~\cite{xu2024,dps}; the right inverse enforces it exactly, by nonlinear back-projection (NLBP) of a chosen generator's proposals, as null-space methods do for linear degradations~\cite{ddnm}. The starting noise sets the prior's influence, allowing perception--distortion choices \emph{after encoding}.

Our adaptation achieves zero PSNR drift through 50 rounds across all reported codec and storage settings, preserving the first reconstructed image exactly in every reported 8-bit chain (Table~\ref{tab:adaptation-summary}). Near $.40$\,bpp under 8-bit storage, adapted HPCM retains $33.95$\,dB at round 50, compared with $31.90$\,dB for MSINN~\cite{msinn2025}. On FFHQ~\cite{ffhq}, NLBP lowers FID from $63.50$ to $7.40$ using a fixed ELIC code, at a PSNR cost of about $2$\,dB. These gains require receiver-side search, taking seconds to tens of seconds per Kodak image.

\vspace{0.2cm}\noindent Our contributions are:
\begin{enumerate}[label=(\roman*),widest=iii,leftmargin=*]
\item We adapt a frozen pretrained codec, without training, so that its encoder and decoder are generalized inverses of one another, quantization included.
\item The adapted codec is exactly idempotent: recompression returns the same image in every tested chain, with an almost zero first-pass loss.
\item Any frozen diffusion prior, including future ones, can also decode the same file, for higher realism and a chosen perception--distortion trade-off. Verified outputs reproduce the code exactly and survive recompression.
\end{enumerate}

\section{Related Work}
\vspace{-0.1cm}
\label{sec:related}
\subsection{Learned image compression}
Learned codecs jointly optimize nonlinear transforms and entropy models to balance bitrate and reconstruction quality~\cite{balle2017,theis2017}. Hyperpriors and spatial/channel context improve probability estimates~\cite{balle2018,minnen2018,charm,checkerboard,elic}; attention, hybrid transforms, and richer context models improve rate--distortion performance~\cite{cheng2020,tcm,hpcm,mlicpp}. Variable-rate coding~\cite{song2021sft} and CompressAI~\cite{compressai} broaden the available backbones. Our adaptation preserves their neural weights while handling quantization and context dependencies.

\subsection{Idempotency and inverse structure}
Successive-compression drift~\cite{instability,kim2022successive} motivates flow and invertible codecs~\cite{flowcodec,xie2021invertible,cai2022iat,cai2024i2c,msinn2025} and coherent demodulation~\cite{stablecdt}. Li et al.~\cite{rightinverse} establish right-invertibility as a route to idempotency and train a codec with pseudoinverse layers and learned null-space components. Our adaptation leaves existing codec weights frozen. IGN~\cite{ign} instead learns idempotent generative projections. JPEG reconstruction~\cite{li2026jpegtv}, changing-quality compression~\cite{strongidempotence2023}, and latent quality conversion~\cite{cai2026hcf} address complementary settings.

SPNN~\cite{spnn} supplies the inverse/completion structure and NLBP; the training-free wrapper of~\cite{dejavae} instantiates it for continuous transforms, without quantization, entropy coding or recompression. We build the right inverse of the quantized codec. Linear null-space methods~\cite{ddrm,ddnm} project exactly and nonlinear posterior guidance~\cite{dps} approximately; NLBP projects exactly through a nonlinear encoder.

\subsection{Generative compression and pretrained priors}
Adversarial, diffusion and foundation-model codecs improve perceptual quality~\cite{hific,msillm,cdc,marimilani,perco,diffeic,dlf,oscar,codlite,aeic,stablecodec}. Coding refinements~\cite{peng2026perq,xu2025picd,liang2025svi} and perception--distortion control~\cite{latenttradeoff,brenig2025,rdptradeoff} expand decoding choices.

Training-free methods combine pretrained components: Xu et al.~\cite{xu2024} use idempotence-constrained inversion; Fast training-free perceptual compression~\cite{fast2025} uses noise-start decoding, following the principle of SDEdit~\cite{sdedit}; rate--distortion guidance~\cite{jiang2025rddm} and DCIC~\cite{dcic2026} constrain diffusion reconstructions. We replace approximate guidance with exact NLBP projection onto a verified target for both streams.

\section{Preliminaries}
\vspace{-0.1cm}
\label{sec:prelim}

\paragraph{Codec representations.}
Let $E$ and $D$ denote the analysis and synthesis transforms of a pretrained codec; each may comprise several pretrained components. The encoder maps an image $x$ to a continuous latent $y=E(x)$. In a hyperprior codec, quantization yields two transmitted symbol sets:
\vspace{-0.1cm}
\begin{equation}
s_z=Q_z(h_a(y)),\qquad s_y=Q_y(y;\hat z),
\end{equation}
where $h_a$ is the hyper-analysis transform and $\hat z$ is reconstructed from $s_z$. Entropy decoding first recovers the hyper symbols and then the main symbols, using parameters predicted from $\hat z$ and, where applicable, previously decoded symbols~\cite{balle2018,minnen2018}. For scalar mean-conditioned quantization with offset $\mu$ and step $\delta_q$,
\vspace{-0.1cm}
\begin{equation}
s_y=\round{(y-\mu)/\delta_q},\qquad \hat y=\mu+\delta_q s_y,
\label{eq:meanquant}
\end{equation}
elementwise, with offsets and steps that may depend on the hyperlatent and decoded context. The compressed file packs the entropy-coded streams with decoding metadata. Without an entropy coder, we evaluate symbol preservation and estimated rate instead of file equality.

\paragraph{Consistency.}
We distinguish continuous-latent recovery, symbol equality, compressed-file equality, and reconstructed-image equality. Reproducing the symbols gives the same file when entropy coding and metadata are deterministic and unchanged; a fixed file gives the same image when decoder settings and randomness are also fixed. Image equality is evaluated at the next encoder's input: floating-point values or decoded 8-bit pixels. Code preservation need not recover the original continuous latent.

\subsection{Right inverses and the pretrained wrapper}
\label{sec:wrapper}
A right inverse reconstructs an image mapping to a specified latent. It need not recover the original image, since several images may share that latent. A map $g:\mathbb R^n\to\mathbb R^m$ is surjective if every target has a preimage. SPNNs~\cite{spnn} provide dimension-reducing maps with explicit right inverses through a bijective completion
\vspace{-0.1cm}
\begin{equation}
G(x)=(g(x),q(x)),
\end{equation}
where $q(x)$ contains the complementary coordinates needed to recover $x$ once $g(x)$ is fixed. The pseudo-inverse induced by this completion is
\vspace{-0.1cm}
\begin{equation}
g^\dagger(v)=G^{-1}(v,0),\qquad gg^\dagger=\mathrm{Id}.
\end{equation}
Its choice depends on $G$; unlike the linear Moore--Penrose inverse, it does not generally minimize pixel-space distance. Appendix~\ref{sec:spnn-prelim} gives the construction and identities.

Following the training-free wrapper of~\cite{dejavae}, we instantiate this construction using frozen pretrained transforms. An orthonormal patch-PCA basis splits centered pixels into
\vspace{-0.1cm}
\begin{equation}
a(x)=U_0(x-\mu),\qquad b(x)=U_1(x-\mu).
\end{equation}
Here $\mu$ denotes the PCA mean, distinct from the quantization offset in Eq.~\eqref{eq:meanquant}. The rows of $U_0$ and $U_1$ together form a complete orthonormal basis. The low-variance coordinates $a(x)$ have the dimension of the latent; $b(x)$ contains the rest. Define
\vspace{-0.1cm}
\begin{align}
J(a,b)&=\mu+U_0^\top a+U_1^\top b,\nonumber\\
P_1(x)&=J(0,b(x)),\qquad f(b)=E(J(0,b)).
\end{align}
For gain $c>0$, the encoder and latent write are
\vspace{-0.1cm}
\begin{align}
\enc(x)&=f(b(x))+c\,a(x),\label{eq:encoder}\\
W(v,b)&=J\!\left(\frac{v-f(b)}c,b\right).\label{eq:write}
\end{align}
The write operation changes the low-variance coordinates so that the wrapped encoder returns the target $v$, while preserving the selected complementary content:
\vspace{-0.1cm}
\begin{equation}
\enc(W(v,b))=v.
\end{equation}
The frozen decoder supplies complementary content through $r(v)=b(D(v))$, giving the right inverse
\vspace{-0.1cm}
\begin{equation}
\dec(v)=W(v,r(v)),\qquad \enc\dec=\Id.
\label{eq:decoder}
\end{equation}
These identities hold in real arithmetic and require compatible dimensions and spatial alignment. PCA is fitted once on calibration data; training-free means that no neural weights are optimized. The gain $c$ affects reconstruction fidelity and numerical robustness, although the exact-write identity holds for every $c>0$. Appendix~\ref{sec:gain-fidelity} gives the deviation analysis. Neither the choice of low-variance coordinates nor the identity alone guarantees a small perceptual change. An external clamp can break the identity.

\begin{figure*}[!t]
\centering
\resizebox{\textwidth}{!}{%
\definecolor{ccInk}{HTML}{303030}
\definecolor{ccMuted}{HTML}{707070}
\definecolor{ccGray}{HTML}{EFEFEF}
\definecolor{ccBlue}{HTML}{315D80}
\definecolor{ccLight}{HTML}{E6EDF3}
\begin{tikzpicture}[x=1cm,y=1cm,
 font=\sffamily\fontsize{14}{16}\selectfont,
 block/.style={rounded corners=2pt,draw=none,fill=ccGray,text=ccInk,
  minimum width=1cm,minimum height=.8cm,align=center,inner sep=5pt},
 frozen/.style={block,fill=ccBlue,text=white,font=\sffamily\fontsize{16}{18}\selectfont},
 tensor/.style={draw=none,text=ccInk,inner sep=0pt},
 op/.style={circle,draw=ccInk,line width=.9pt,fill=white,
  minimum size=.6cm,inner sep=0pt,font=\sffamily\fontsize{18}{20}\selectfont},
 flow/.style={-{Stealth[length=2.4mm,width=1.6mm]},draw=ccInk!75,line width=.9pt,shorten >=1pt},
 context/.style={flow,densely dashed,draw=ccBlue!80},
 fork/.style={circle,fill=ccInk!75,draw=none,minimum size=3.5pt,inner sep=0pt},
 label/.style={font=\sffamily\fontsize{13}{15}\selectfont,text=ccMuted},
 title/.style={anchor=west,text=ccInk,font=\sffamily\bfseries\fontsize{14}{16}\selectfont}]
\path[use as bounding box] (0,.7) rectangle (25.15,11.4);
\newcommand{\splitTensor}[5]{%
 \node[minimum width=1.2cm,minimum height=1.3cm,inner sep=0pt] (#1) at (#2,#3) {};
 \path[fill=ccGray,rounded corners=1pt] (#2-.6,#3-.65) rectangle (#2+.6,#3+.65);
 \draw[white,line width=1.5pt] (#2-.6,#3+.15)--(#2+.6,#3+.15);
 \node[tensor] (#1zero) at (#2,#3+.4) {$#4$};
 \node[tensor] (#1one) at (#2,#3-.25) {$#5$};
 \coordinate (#1zeroeast) at (#2+.6,#3+.4);
 \coordinate (#1oneeast) at (#2+.6,#3-.25);
 \coordinate (#1onewest) at (#2-.6,#3-.25);
 \coordinate (#1zeronorth) at (#2,#3+.65);
}

\node[title] at (.05,11) {(a) Encoding: adapted transform, quantization, and entropy coding};
\node[anchor=east,label] at (24.5,11) {\textcolor{ccBlue}{\rule{8pt}{7pt}}\; Frozen network};
\node[anchor=west,label] at (.25,10.2) {Adapted $E'$};

\node[tensor,font=\fontsize{17}{19}\selectfont] (x) at (.3,8.9) {$x$};
\node[block] (pca) at (1.8,8.9) {$\mathrm{PCA}$};
\splitTensor{xs}{3.8}{8.9}{a}{b}
\node[block,font=\fontsize{16}{18}\selectfont] (gain) at (10.1,9.55) {$c$};
\splitTensor{ezero}{6}{8.3}{0}{b}
\node[block,minimum width=1.6cm] (ipca) at (8,8.3) {$\mathrm{PCA}^{-1}$};
\node[frozen] (e) at (10.1,8.3) {$E$};
\node[op] (sum) at (14.5,8.9) {$+$};
\node[tensor,font=\fontsize{17}{19}\selectfont] (y) at (18.2,8.9) {$y$};
\draw[flow] (x.east)--(pca.west);
\draw[flow] (pca.east)--(xs.west);
\draw[flow] (xszeroeast)--(4.85,9.3)--(4.85,9.55)--(gain.west);
\draw[flow] (gain.east)--(14.5,9.55)--(sum.north);
\draw[flow] (xsoneeast)--(4.85,8.65)--(4.85,8.05)--(ezeroonewest);
\draw[flow] (ezero.east)--(ipca.west);
\draw[flow] (ipca.east)--(e.west);
\draw[flow] (e.east)--(14.5,8.3)--(sum.south);
\draw[flow] (sum.east)--(y.west);

\node[block,fill=ccLight] (qy) at (18.2,7.65) {$Q_y$};
\node[block] (ecy) at (18.2,6.25) {EC};
\node[draw=ccBlue!60,rounded corners=1pt,fill=white,line width=.7pt,
 minimum width=1.45cm,minimum height=.54cm,inner sep=0pt] (by) at (18.2,5.05) {};
\foreach \i/\bit in {0/0,1/1,2/0,3/0,4/1,5/1,6/0,7/1}{
 \pgfmathsetmacro{\xx}{18.2-.68+.17*\i}
 \ifnum\bit=1
  \fill[ccInk] (\xx,4.84) rectangle (\xx+.17,5.26);
  \node[text=white,font=\ttfamily\fontsize{8}{8}\selectfont,inner sep=0pt] at (\xx+.085,5.05) {1};
 \else
  \node[text=ccInk,font=\ttfamily\fontsize{8}{8}\selectfont,inner sep=0pt] at (\xx+.085,5.05) {0};
 \fi
}
\node[label,anchor=west,font=\fontsize{13}{15}\selectfont] at (19,5.05) {$B_y$};
\node[block] (edy) at (18.2,3.85) {ED};
\node[tensor,font=\fontsize{17}{19}\selectfont] (yh) at (18.2,2.8) {$\hat y$};
\node[frozen,minimum width=1.2cm] (ha) at (21.3,8.9) {$h_a$};
\node[block,fill=ccLight] (qz) at (23.8,7.65) {$Q_z$};
\node[block] (ecz) at (23.8,6.25) {EC};
\node[draw=ccBlue!60,rounded corners=1pt,fill=white,line width=.7pt,
 minimum width=1.45cm,minimum height=.54cm,inner sep=0pt] (bz) at (23.8,5.05) {};
\foreach \i/\bit in {0/1,1/0,2/1,3/1,4/0,5/0,6/1,7/0}{
 \pgfmathsetmacro{\xx}{23.8-.68+.17*\i}
 \ifnum\bit=1
  \fill[ccInk] (\xx,4.84) rectangle (\xx+.17,5.26);
  \node[text=white,font=\ttfamily\fontsize{8}{8}\selectfont,inner sep=0pt] at (\xx+.085,5.05) {1};
 \else
  \node[text=ccInk,font=\ttfamily\fontsize{8}{8}\selectfont,inner sep=0pt] at (\xx+.085,5.05) {0};
 \fi
}
\node[label,anchor=west,font=\fontsize{13}{15}\selectfont] at (24.58,5.05) {$B_z$};
\node[block] (edz) at (23.8,3.85) {ED};
\node[tensor,font=\fontsize{17}{19}\selectfont] (zh) at (23.8,2.8) {$\hat z$};
\node[frozen,minimum width=1.2cm] (hs) at (21.3,2.8) {$h_s$};
\draw[flow] (y.south)--(qy.north);
\draw[flow] (qy.south)--(ecy.north);
\draw[flow] (ecy.south)--(by.north);
\draw[flow] (by.south)--(edy.north);
\draw[flow] (edy.south)--(yh.north);
\draw[flow] (ha.east)--(23.8,8.9)--(qz.north);
\draw[flow] (qz.south)--(ecz.north);
\draw[flow] (ecz.south)--(bz.north);
\draw[flow] (bz.south)--(edz.north);
\draw[flow] (edz.south)--(zh.north);
\draw[flow] (zh.west)--(hs.east);
\node[label,anchor=east] at (17.75,6.95) {$s_y$};
\node[label,anchor=west] at (24.25,6.95) {$s_z$};

\coordinate (params) at (19.9,2.8);
\draw[context] (hs.west)--(params)--(19.9,3.85)--(edy.east);
\draw[context] (params)--(19.9,6.25)--(ecy.east);
\draw[context] (params)--(19.9,7.65)--(qy.east);
\node[label,anchor=south,text=ccBlue,fill=white,inner sep=2pt] at (18.95,7.68) {$\mu,\delta_q$};

\draw[flow] (y.east)--(ha.west);

\node[title] at (.05,6.3) {(b) Decoding and latent write};
\node[anchor=west,label] at (.25,5.45) {Explicit write $W$};
\begin{scope}[xshift=17.2cm,xscale=-.86]
\node[tensor,font=\fontsize{17}{19}\selectfont] (dy) at (.3,2.8) {};
\node[frozen] (d) at (1.65,2.8) {$D$};
\node[block] (dpca) at (3.45,2.8) {$\mathrm{PCA}$};
\splitTensor{ds}{5.4}{2.8}{a_D}{b_D}
\splitTensor{dzero}{7.15}{4.3}{0}{b_D}
\node[block,minimum width=1.6cm] (dipca) at (9.3,4.3) {$\mathrm{PCA}^{-1}$};
\node[frozen] (de) at (11.45,4.3) {$E$};
\node[block,minimum width=1.7cm] (minus) at (13.9,4.3) {$y^\star-E(\cdot)$};
\node[block,font=\fontsize{16}{18}\selectfont] (inv) at (16.55,4.3) {$1/c$};
\splitTensor{out}{15.75}{2.8}{a^+}{b_D}
\node[block,minimum width=1.6cm] (outpca) at (17.85,2.8) {$\mathrm{PCA}^{-1}$};
\node[tensor,font=\fontsize{17}{19}\selectfont] (xhat) at (19.45,2.8) {$\tilde x$};

\node[fork] (bfork) at (6.25,2.55) {};
\draw[flow] (d.west)--(dpca.east);
\draw[flow] (dpca.west)--(ds.east);
\draw[flow] (dsoneeast)--(outonewest);
\draw[flow] (bfork.north)--(6.25,4.05)--(dzeroonewest);
\draw[flow] (dzero.west)--(dipca.east);
\draw[flow] (dipca.west)--(de.east);
\draw[flow] (de.west)--(minus.east);

\draw[flow] (minus.west)--(inv.east);
\draw[flow] (inv.south)--(16.55,3.7)-|(outzeronorth);
\draw[flow] (out.west)--(outpca.east);
\draw[flow] (outpca.west)--(xhat.east);
\node[label] at (10.8,3.05) {preserve $b_D$};
\end{scope}

\draw[flow] (yh.west)--(d.east);

\node[block,
 minimum width=4.1cm,minimum height=1.45cm,inner sep=0pt] (target) at (14.4,5.1) {};
\node[font=\sffamily\fontsize{11}{13}\selectfont,text=ccInk] at (14.4,5.47) {Target search};
\node[frozen,minimum width=.66cm,minimum height=.5cm,inner sep=2pt,font=\fontsize{13}{15}\selectfont] at (13.05,4.86) {$E$};
\node[frozen,minimum width=.66cm,minimum height=.5cm,inner sep=2pt,font=\fontsize{13}{15}\selectfont] at (13.95,4.86) {$h_a$};
\node[block,minimum width=.66cm,minimum height=.5cm,inner sep=2pt,font=\fontsize{13}{15}\selectfont] at (14.85,4.86) {$Q_y$};
\node[block,minimum width=.66cm,minimum height=.5cm,inner sep=2pt,font=\fontsize{13}{15}\selectfont] at (15.75,4.86) {$Q_z$};

\draw[flow] (target.west)--(minus.north |- target.west)--(minus.north);
\node[label,anchor=south,fill=white,inner sep=2pt] at (10.6,5.12) {$y^\star$};

\node[fork] (ysearchfork) at (17.25,2.8) {};
\draw[flow] (ysearchfork.north)--(17.25,5.46)--(16.45,5.46);
\node[label,anchor=south,fill=white,inner sep=1pt] at (16.98,5.48) {$\hat y$};

\draw[draw=ccBlue,line width=.9pt,rounded corners=2pt]
 (zh.south)--(23.8,1.08)--(16.88,1.08)--(16.88,2.6);
\draw[draw=ccBlue,line width=.9pt,preaction={draw=white,line width=4pt}]
 (16.88,2.6) arc[start angle=-90,end angle=90,radius=.2];
\draw[flow,draw=ccBlue,rounded corners=2pt]
 (16.88,3)--(16.88,4.65)--(16.45,4.65);
\node[label,text=ccBlue,anchor=north,fill=white,inner sep=1pt] at (16.68,4.58) {$\hat z$};
\end{tikzpicture}
}
\par\vspace{-0.2cm}\caption{\small \textbf{Floating-point recompression with frozen networks.} Target search selects $y^\star$; $W$ writes it while preserving $b_D$. The first encode uses $E$, subsequent ones $E'$. Here $a_D=a(D(\hat y))$, $b_D=b(D(\hat y))$, $a^+=(y^\star-f(b_D))/c$, and $\mathrm{PCA}^{-1}=J$ (Sec.~\ref{sec:wrapper}). $h_s$ is hyper-synthesis; dashed arrows indicate conditioning, with context networks implicit. $B_y,B_z$ denote logical main/hyper bitstreams; EC/ED denote entropy encoding/decoding, and ED includes latent reconstruction. The coding path is representative; GLC stores uncoded symbols and reports estimated rate.}
\label{fig:adaptation}
\end{figure*}
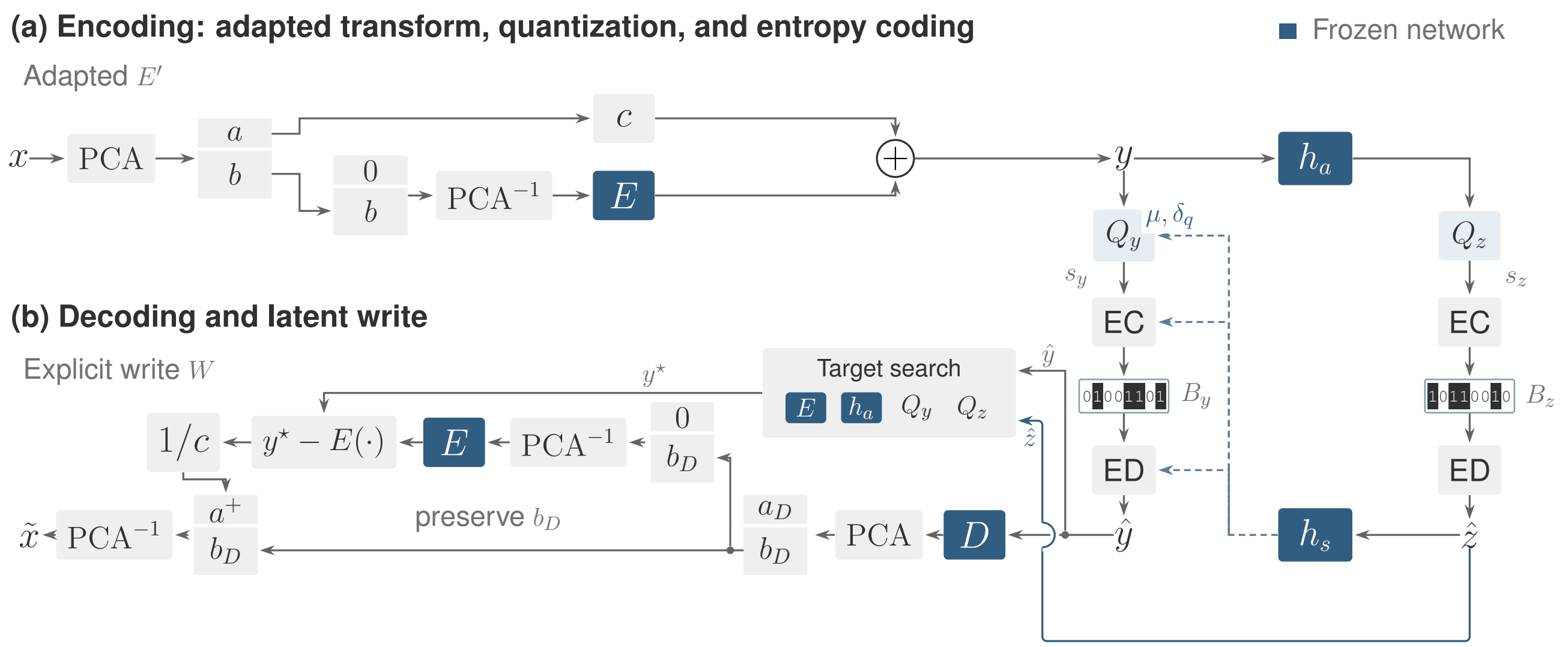

\subsection{Back-projection of image proposals}
\label{sec:backprojection-prelim}
Back-projection corrects a candidate image to agree with given measurements. For linear $A$, candidate $u$, and feasible measurements $v\in\operatorname{range}(A)$, the closest consistent image is
\vspace{-0.1cm}
\begin{align}
u^+&=\operatorname*{arg\,min}_{x:\,Ax=v}\|x-u\|_2^2\nonumber\\
&=u+A^\dagger(v-Au),
\end{align}
where $A^\dagger$ is the Moore--Penrose pseudo-inverse. This enforces $Au^+=v$ while preserving its null-space component.

The wrapper extends this nonlinearly. Its completion is $G(x)=(\enc(x),b(x))$, with $G^{-1}(v,b)=W(v,b)$. With $g=\enc$ and completion-induced inverse $g^\dagger(v)=G^{-1}(v,0)$ in NLBP~\cite{spnn}, we replace latent coordinates:
\vspace{-0.1cm}
\begin{equation}
u^+=W(v,b(u)).
\end{equation}
This update satisfies $\enc(u^+)=v$ while retaining the proposal's complementary coordinates. It is closest in completion coordinates, generally not in Euclidean distance.

A generative model can supply the proposal $u$, and the correction enforces the latent constraint (general NLBP: Appendix~\ref{sec:nlbp-prelim}; our corrections: Sec.~\ref{sec:nlbp}). These identities concern continuous latents in real arithmetic; preserving a compressed file also requires reproducing its quantized symbols and entropy-coding information, including after 8-bit storage, which Sec.~\ref{sec:method} addresses.

\section{Method}
\vspace{-0.1cm}
\label{sec:method}
\subsection{Adapting the wrapper to quantized codecs}
\label{sec:side}
Our adaptation preserves a discrete representation using a continuous right inverse (Fig.~\ref{fig:adaptation}). The receiver selects a target $y^\star$ that reproduces both received symbol sets, writes it with Eq.~\eqref{eq:write}, and verifies re-encoding. The write retains complementary content from $D(\hat y)$; all neural weights remain frozen. Initial encoding uses the original $E$, retaining its file and rate, while recompression uses $E'$. This also supports files encoded before adaptation. Each codec supplies its quantizers and entropy checks. Receiver-side repair handles 8-bit storage (Sec.~\ref{sec:formats}) and clamping (Appendix~\ref{sec:clamp-repair}).

\paragraph{Why the wrapper alone is not enough.}
The wrapper writes any continuous target into an image so that $E'$ returns it exactly. The natural target, the dequantized latent $\hat y$, is not enough: the hyper symbols were computed from the continuous $y$, and re-encoding an image that carries $\hat y$ computes $h_a(\hat y)$ instead, which can change the hyper symbols. In mean-conditioned codecs, this can also change the main quantization offsets and, consequently, the main symbols. Appendix~\ref{app:toy-example} gives a scalar example.

\paragraph{The feasible set.}
We seek a target $t$ that reproduces both received symbol streams, $s_y$ and $s_z$, when re-quantized. The set of all such targets is
\vspace{-0.1cm}
\begin{equation}
\label{eq:cell}
\mathcal F(s)=
\underbrace{\{t:Q_y(t;\hat z)=s_y\}}_{\mathcal C_y(s)}
\cap
\underbrace{\{t:Q_z(h_a(t))=s_z\}}_{\mathcal C_z(s)}.
\end{equation}
$\mathcal F(s)$ is never empty (Appendix~\ref{app:feasible-nonempty}), but the decoder does not know a member and must search. For numerical robustness, it searches the margin-constrained set $\mathcal F_m(s)$, keeping margins $m_y$ and $m_z$ inside the main and hyper cells.

\begin{figure}[!t]
\centering\includegraphics[width=\linewidth]{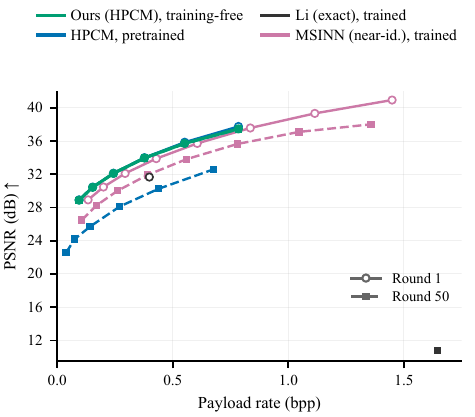}
\par\vspace{-0.1cm}\caption{\small\textbf{Quality retained through 8-bit image recompression.} Mean Kodak PSNR versus payload rate at rounds 1 (circles) and 50 (squares); round-50 points describe recompressed files, not operating points. Ours (HPCM, in-memory chain) keeps its round-1 image at all six rates (rate drift $\le.0005$\,bpp); Li's released model is evaluated outside its native float chain.}
\label{fig:idem-rd}
\end{figure}

\paragraph{Decoder-side selection.}
For the aligned reconstruction $x_{\rm raw}=D(\hat y)$, the write changes only the low-variance coordinates, so for every target $t$
\vspace{-0.1cm}
\begin{equation}
\|W(t,b(x_{\rm raw}))-x_{\rm raw}\|_2=\|t-E'(x_{\rm raw})\|_2/c,
\label{eq:closest}
\end{equation}
and the closest consistent reconstruction corresponds to
\vspace{-0.1cm}
\begin{equation}
y^\star=\operatorname*{arg\,min}_{t\in\mathcal F_m(s)}\|t-E'(x_{\rm raw})\|_2,
\label{eq:select}
\end{equation}
the latent-space counterpart of back-projection. Eq.~\eqref{eq:select} motivates a feasibility-first search over the main and hyper cells rather than an exact minimization. A target is accepted only when the codec's own quantizers return $s$; if none is accepted, the decoder writes $\hat y$, which preserves the code only when $\hat y$ is itself feasible. The search uses only the received symbols, frozen copies of $E$ and $h_a$, and a single synthesis output; it does not guarantee the minimizer of Eq.~\eqref{eq:select}. The search steps, losses, guards, restarts and per-codec settings are in Appendices~\ref{sec:target-search} and~\ref{sec:implementation-settings}.

\paragraph{What is guaranteed.}
Let $\tilde x=W(y^\star,b(x_{\rm raw}))$, and let $\Sigma$ map a latent to its symbol record. In real arithmetic $E'(\tilde x)=y^\star$, so
\vspace{-0.1cm}
\begin{equation}
\label{eq:search}
E'(\tilde x)\in\mathcal F(s)\quad\Longrightarrow\quad\Sigma(E'(\tilde x))=s.
\end{equation}
In floating point, $E'(\tilde x)=y^\star+\varepsilon$; the main symbols are kept whenever $\|\varepsilon\|_\infty<m_y$, and the hyper symbols whenever the induced change of $h_a$ stays below $m_z$. With deterministic decoding, receiving $s$ again reproduces $\tilde x$, so every later generation sends $s$ and reconstructs $\tilde x$; equal bytes additionally require deterministic entropy coding and identical metadata. This argument covers the floating-point chain on the tested implementation; it is conditional on an accepted target and does not establish cross-platform equality. Writing follows the codec's output clamp, so $\tilde x$ may leave $[0,1]$; clamping or 8-bit storage then moves it after the check.

\subsection{Consistency under 8-bit storage}
\label{sec:formats}
An 8-bit recompression chain passes a clipped and rounded image to the next encoder. Let $S_8$ denote clipping to $[0,1]$, rounding to 8-bit RGB (ties to even), a lossless storage round trip, and conversion back to the codec's input normalization (Appendix~\ref{sec:recompression-protocols}). Write $\mathcal G_8=\operatorname{range}(S_8)$. The exact write gives $E'(\tilde x)\in\mathcal F(s)$, but not $E'(S_8(\tilde x))\in\mathcal F(s)$: storage changes both coordinate groups and can cross main or hyper cell boundaries. The receiver therefore seeks an image in
\vspace{-0.1cm}
\begin{equation}
 \mathcal P_8(s)=\{x\in\mathcal G_8:E'(x)\in\mathcal F(s)\}.
 \label{eq:png-feasible}
\end{equation}
Continuous feasibility permits $\mathcal P_8(s)=\varnothing$ (Appendix~\ref{sec:png-nonexistence}).

\paragraph{Decoder-side repair.}
Starting from $y^\star$, a closed loop corrects the write target,
\vspace{-0.1cm}
\begin{equation}
 t_0=y^\star,\qquad x_k=S_8\!\left(W(t_k,b(x_{\rm raw}))\right),
 \label{eq:png-repair}
\end{equation}
re-encodes each trial, and accepts it if $\Sigma(E'(x_k))=s$; otherwise main-cell overshoots and a hyper-cell gradient correction update $t_k$. When needed, lattice descent then moves individual pixel channels by one grey level. The target is a control variable; feasibility is tested on the stored image. If a candidate fails this test, the decoder repeats the search with a larger budget (a doubled update cap, a wider hyper-cell guard, then a projected search inside the pixel box; Appendix~\ref{app:escalation}); only if every stage fails does it return a best-effort image without claiming consistency.

\paragraph{Image stability versus file preservation.}
When the received symbols cannot be reproduced on the 8-bit grid, an optional stabilization stage may accept a different symbol record (mechanics in Appendix~\ref{sec:png-details}; configurations in Table~\ref{tab:png-settings}). On success, the 8-bit decoder $D_8$ satisfies
\vspace{-0.1cm}
\begin{equation}
 D_8(\Sigma(E'(x^\star)))=x^\star,
 \qquad x^\star=D_8(s_0).
 \label{eq:png-fixed}
\end{equation}
The image is then fixed from the first reconstruction, but if the accepted symbols differ from $s_0$, the file changes once, at generation two, and stays fixed thereafter. We therefore report image equality and first-file equality separately. Neither search success nor convergence of the outer loop is guaranteed; an exhausted budget is not a certificate.

\begin{table*}[!t]
\caption{\small\textbf{Adaptation across codecs: coverage, first-pass cost, and drift.}
All our variants keep neural weights frozen. Rates is the number of tested compression settings. Exactness counts all tested images across all compression rates. Float is unclamped, and Clipped float is clamped. The PSNR columns (dB) give lower bounds on the worst (minimum) rate-wise image-mean change: $\Delta$PSNR$_1$ is adapted minus original decoding of the same first file; drift is last minus first round. BPP drift retains its signed bounds. Image / First file denotes equality to the first image / initial compressed file at every checked round; Base is the original codec on the same chain. Li and MSINN require training; Li's native chain is Float. Xu is a training-free ELIC decoder ($\lambda=.45$) on $256^2$ crops.
$^{\dagger}$File changes once, at round 2; image kept.
$^{\ast}$Development set: preset search gives 493/500 (DIV2K) and 141/144 (ELIC).
$^{\ddagger}\lambda\le.013$.
$^{\P}$Symbol level, estimated rate. Details: Appendix~\ref{sec:serialization-results}.}\par\vspace{-0.1cm}
\label{tab:adaptation-summary}
\centering\small
\setlength{\tabcolsep}{2pt}
\renewcommand{\arraystretch}{1.08}
\begin{tabular*}{\textwidth}{@{\extracolsep{\fill}}cllrrrrrrrr@{}}
\toprule
 & Codec / dataset & Chain & Rates & Rounds & \shortstack{Worst\\$\Delta$PSNR$_1\geq$} & \shortstack{Worst\\PSNR drift $\geq$} & \shortstack{BPP\\drift} & \shortstack{Base\\exact} & \shortstack{First file\\exact} & \shortstack{Image\\exact} \\
\midrule
\multirow{18}{*}{\rotatebox{90}{\textit{Ours}}} & AEIC~\cite{aeic} / Kodak & 8-bit & 5 & 50 & $-0.001$ & 0 & 0 & 0/120 & 120/120 & 120/120 \\
 & HPCM~\cite{hpcm} / Kodak & 8-bit & 6 & 50 & $-0.290$ & 0 & $\leq.0005$ & 0/144 & 137/144$^{\dagger}$ & 144/144 \\
 & ELIC~\cite{elic} / Kodak & 8-bit & 6 & 50 & $-0.085$ & 0 & 0 & 0/144 & 144/144$^{\ast}$ & 144/144$^{\ast}$ \\
 & MS-ILLM~\cite{msillm} / Kodak & 8-bit & 5 & 50 & $-0.007$ & 0 & 0 & 0/120 & 120/120 & 120/120 \\
 & Hyperprior~\cite{balle2018} / Kodak (q1--q7) & 8-bit & 7 & 50 & $-1.180$ & 0 & 0 & 0/168 & 168/168 & 168/168 \\
 & TCM~\cite{tcm} / Kodak & 8-bit & 6 & 50 & $-0.047$ & 0 & 0 & 0/144 & 144/144 & 144/144 \\
 & Mean-scale~\cite{minnen2018} / Kodak (q1--q7) & 8-bit & 7 & 50 & $-0.806$ & 0 & 0 & 0/168 & 168/168 & 168/168 \\
 & AEIC~\cite{aeic} / Kodak & Float & 5 & 50 & $-0.0024$ & 0 & 0 & 0/120 & 120/120 & 120/120 \\
 & HPCM~\cite{hpcm} / Kodak & Float & 6 & 50 & $-0.138$ & 0 & 0 & 0/144 & 144/144 & 144/144 \\
 & AEIC~\cite{aeic} / Kodak & Clipped float & 5 & 50 & $-0.001$ & 0 & 0 & 0/120 & 120/120 & 120/120 \\
 & HPCM~\cite{hpcm} / Kodak & Clipped float & 4$^{\ddagger}$ & 50 & $-0.041$ & 0 & 0 & 0/96 & 96/96 & 96/96 \\
 & ELIC~\cite{elic} / Kodak & Float & 6 & 50 & $-0.026$ & 0 & 0 & 0/144 & 144/144 & 144/144 \\
 & GLC~\cite{glc2024} / Kodak & Float & 4 & 50 & $+0.007$ & 0 & 0 & 0/96$^{\P}$ & n/a & 96/96 \\
 & AEIC~\cite{aeic} / DIV2K (100) & Float & 5 & 50 & $-0.040$ & 0 & 0 & 0/500 & 500/500 & 500/500 \\
 & AEIC~\cite{aeic} / DIV2K (100) & 8-bit & 5 & 50 & $-0.001$ & 0 & 0 & 0/500 & 500/500$^{\ast}$ & 500/500$^{\ast}$ \\
 & AEIC~\cite{aeic} / CLIC (41) & 8-bit & 5 & 50 & $-0.001$ & 0 & 0 & 0/205 & 205/205 & 205/205 \\
 & ELIC~\cite{elic} / Kodak ($256^2$) & 8-bit & 1 & 50 & $-0.053$ & 0 & 0 & 0/24 & 24/24 & 24/24 \\
 & ELIC~\cite{elic} / CLIC (41, $256^2$) & 8-bit & 1 & 50 & $-0.081$ & 0 & 0 & 0/41 & 41/41 & 41/41 \\
\midrule
\multirow{7}{*}{\rotatebox{90}{\shortstack{\textit{Other consistency}\\\textit{methods}}}} & Li (exact)~\cite{rightinverse} / Kodak & 8-bit & 1 & 50 & -- & $-20.91$ & $+1.245$ & -- & 0/24 & 0/24 \\
 & MSINN~\cite{msinn2025} / Kodak & 8-bit & 8 & 50 & -- & $-2.93$ & $\geq-0.092$ & -- & 0/192 & 0/192 \\
 & Li (exact)~\cite{rightinverse} / Kodak & Float & 1 & 50 & -- & 0 & $+0.007$ & -- & 0/24$^{\dagger}$ & 24/24 \\
 & Li (exact)~\cite{rightinverse} / Kodak & Clipped float & 1 & 50 & -- & $-15.56$ & $+0.440$ & -- & 0/24 & 0/24 \\
 & MSINN~\cite{msinn2025} / Kodak & Clipped float & 8 & 50 & -- & $-2.36$ & $\geq-0.078$ & -- & 0/192 & 0/192 \\
 & Xu~\cite{xu2024} / Kodak ($256^2$) & 8-bit & 1 & 50 & $-2.79$ & $-8.16$ & $-0.130$ & 0/24 & 0/24 & 0/24 \\
 & Xu~\cite{xu2024} / CLIC (41, $256^2$) & 8-bit & 1 & 50 & $-2.48$ & $-7.73$ & $-0.101$ & 0/41 & 0/41 & 0/41 \\
\bottomrule
\end{tabular*}
\end{table*}

\begin{table}[!t]
\caption{\small\textbf{Quality under 8-bit image recompression.} Rounds 1 and 50 on Kodak (24 images), starting from the same first file. MS-S.: MS-SSIM; KID is multiplied by $10^3$. All values are measured; BPP is payload. Chains use the 8-bit grid. FID/KID use 2,304 $64^2$ patches. \textbf{Bold}: better score within codec and round. CLIC and DIV2K: Appendix Table~\ref{tab:grand-all}.}\par\vspace{-0.1cm}
\label{tab:recompression-main}
\centering\footnotesize
\setlength{\tabcolsep}{0.65pt}
\renewcommand{\arraystretch}{1.10}
\begin{tabular*}{\linewidth}{@{\extracolsep{\fill}}clrrrrrrr@{}}
\toprule
{\footnotesize Round} & {\footnotesize Decoder} & {\footnotesize BPP} & {\footnotesize PSNR$\uparrow$} & {\footnotesize MS-S.$\uparrow$} & {\footnotesize LPIPS$\downarrow$} & {\footnotesize DISTS$\downarrow$} & {\footnotesize FID$\downarrow$} & {\footnotesize KID$\downarrow$} \\
\midrule
\multicolumn{9}{@{}l}{\textit{AEIC (ft2)}} \\[1pt]
1 & Original & .0379 & \textbf{23.30} & \textbf{.8323} & \textbf{.143} & \textbf{.0820} & \textbf{28.20} & \textbf{1.40} \\
 & Adapted & .0379 & \textbf{23.30} & \textbf{.8323} & \textbf{.143} & \textbf{.0820} & 28.21 & \textbf{1.40} \\
\cmidrule[0.2pt](lr){1-9}
50 & Original & .0552 & 13.32 & .4739 & .638 & .3429 & 137.40 & 68.85 \\
 & Adapted & .0379 & \textbf{23.30} & \textbf{.8323} & \textbf{.143} & \textbf{.0820} & \textbf{28.21} & \textbf{1.40} \\
\midrule
\multicolumn{9}{@{}l}{\textit{HPCM ($\lambda=.013$)}} \\[1pt]
1 & Original & .3778 & \textbf{33.99} & \textbf{.9806} & \textbf{.137} & .1258 & 64.49 & 23.06 \\
 & Adapted & .3778 & 33.95 & \textbf{.9806} & .138 & \textbf{.1254} & \textbf{61.85} & \textbf{20.67} \\
\cmidrule[0.2pt](lr){1-9}
50 & Original & .2713 & 28.12 & .9229 & .261 & .2024 & 93.57 & 36.44 \\
 & Adapted & .3778 & \textbf{33.95} & \textbf{.9806} & \textbf{.138} & \textbf{.1254} & \textbf{61.85} & \textbf{20.67} \\
\bottomrule
\end{tabular*}
\vspace{-0.1cm}
\end{table}

\begin{table}[!t]
\caption{\small\textbf{Comparison with consistency methods on Kodak.} PSNR (dB) after 1, 32 and 50 rounds. HPCM, ours and MSINN: our floating-point-chain measurements (at $.708$\,bpp interpolated between neighbouring rate settings). Li et al.\ and Helminger: published values, partly read from curves, storage protocol unspecified; $^{\dagger}$from the published exact-idempotence claim, not a measured run; --: not reported. Full table: Appendix~\ref{sec:baseline-audit}.}\par\vspace{-0.1cm}
\label{tab:consistency-methods}
\centering\footnotesize
\setlength{\tabcolsep}{3pt}
\begin{tabular}{@{}llrrr@{}}
\toprule
Method & Training & PSNR$_1$ & PSNR$_{32}$ & PSNR$_{50}$ \\
\midrule
\multicolumn{5}{@{}l}{\textit{Initial rate $\approx .708$ bpp}} \\[1pt]
HPCM & pretrained & \textbf{37.16} & 32.75 & 32.20 \\
HPCM + ours & training-free & 37.03 & \textbf{37.03} & \textbf{37.03} \\
MSINN~\cite{msinn2025} & trained & 36.56 & 34.89 & 34.89 \\
Li (exact)~\cite{rightinverse} & trained & $\approx34.50$ & $\approx34.50^{\dagger}$ & $\approx34.50^{\dagger}$ \\
Li (near)~\cite{rightinverse} & trained & 35.79 & 34.92 & 34.92 \\
\midrule
\multicolumn{5}{@{}l}{\textit{Initial rates $.784$ (HPCM) / $.796$ (Helminger) bpp}} \\[1pt]
HPCM & pretrained & \textbf{37.78} & 33.59 & 33.08 \\
HPCM + ours & training-free & 37.64 & \textbf{37.64} & \textbf{37.64} \\
Helminger~\cite{flowcodec,rightinverse} & trained & 32.04 & -- & -- \\
\bottomrule
\end{tabular}
\vspace{-0.3cm}
\end{table}

\subsection{NLBP with independently chosen priors}
\label{sec:nlbp}
Given $s$, the receiver selects $y^\star$ and forms $x_{\mathrm{base}}=W(y^\star,b(D(\hat y)))$. Sampling starts from this reconstruction with added noise; the starting noise controls the prior's influence~\cite{sdedit}. At each step, NLBP~\cite{spnn} corrects the frozen generator's clean-image estimate $u_t$. With $y_t=E'(u_t)$, the update is
\vspace{-0.1cm}
\begin{equation}
\begin{aligned}
v_t^+&=(1-\lambda_t)y_t+\lambda_t y^\star,\\
b_t^+&=b(u_t)+\kappa\lambda_t[r_{\mathrm{ref}}-b(D_{\mathrm{ref}}(y_t))],\\
u_t^+&=W(v_t^+,b_t^+).
\end{aligned}
\label{eq:nlbp-update}
\end{equation}
Here $D_{\mathrm{ref}}$ is the fixed reference decoder, including its configured preprocessing, and $r_{\mathrm{ref}}=b(D_{\mathrm{ref}}(v_{\mathrm{ref}}))$ for a fixed $v_{\mathrm{ref}}\in\{\hat y,y^\star\}$ (Appendix~\ref{sec:nlbp-implementation}). Setting $\kappa=0$ preserves the proposal's complementary coordinates; nonzero $\kappa$ incorporates the reference decoder. Since $E'(u_t^+)=v_t^+$ in real arithmetic, $\lambda_t=1$ enforces the target for either choice. The corrected proposal enters the next sampling step (Appendix Algorithm~\ref{alg:nlbp}); latent priors are corrected in image space. This is generally neither a nearest-pixel projection nor exact posterior sampling.

\paragraph{Final write and verification.}
After sampling, the receiver writes $y^\star$ into the proposal, optionally refines complementary coordinates (Appendix~\ref{sec:nlbp-implementation}), and applies repair for 8-bit chains (Sec.~\ref{sec:formats}). Code preservation requires $\Sigma(E'(x_{\mathrm{out}}))=s$ with the actual encoder and quantizers; failures return a best-effort reconstruction. Randomness seeded from the received symbols makes verified outputs reproducible under recompression with fixed settings.

\begin{figure*}[!t]
  \centering
  \includegraphics[width=0.8\textwidth, trim={0 1.8cm 0 0},clip]{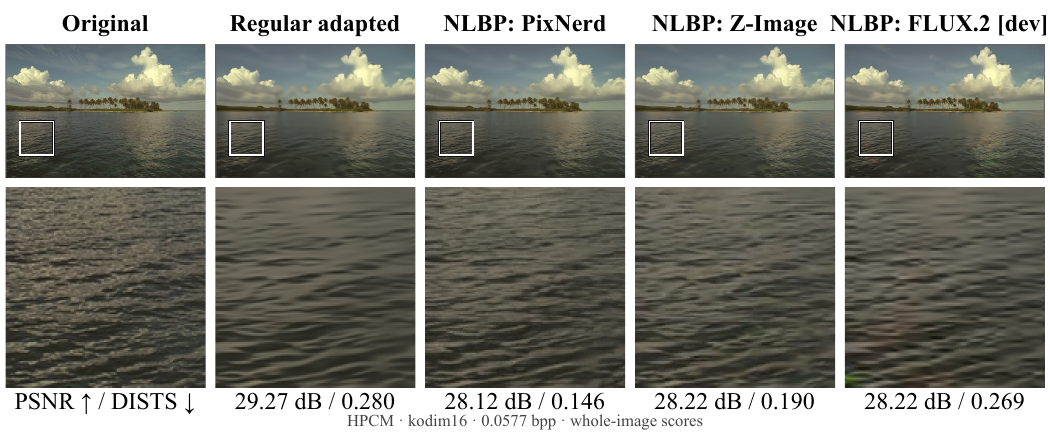}
  \par\vspace{-0.1cm}\caption{\small\textbf{Choosing a pretrained generator at decoding time.}
  All NLBP columns decode the same bitstream.
  Top: full images with the highlighted crop. Bottom: matching
  $128\times128$ details. PixNerd uses its tuned setting; Z-Image and FLUX.2 [dev] use untuned settings.}
  \label{fig:generator-choice}
  \vspace{-0.2cm}
\end{figure*}

\section{Experiments}
\vspace{-0.1cm}
\label{sec:experiments}
We evaluate recompression quality and decoder choice after encoding. Codec and generator weights remain frozen; adaptation uses PCA calibration and inference-time search.

\subsection{Evaluation protocol}
We test unclamped and clamped floating-point chains, and 8-bit image chains as the primary operational test. Table~\ref{tab:adaptation-summary} lists codecs, datasets, rates and rounds. GLC uses the released implementation~\cite{glc2024}; high-resolution AEIC (CLIC Pro and DIV2K validation) retains a padded canvas and the original colour metadata. The first encode uses the original codec. We distinguish first-reconstruction change from later drift, and image equality from compressed-file equality, both relative to round 1; file equality excludes stored-image container bytes. Rates count all entropy-coded streams and transmitted side information, excluding container framing. GLC has no entropy coder: its rates are estimates and equality concerns symbols. We report PSNR, MS-SSIM, LPIPS~\cite{lpips}, DISTS and patch FID/KID (protocols: Appendix~\ref{sec:full-results}).

\begin{table}[!t]
\caption{\small\textbf{Ablations} (Kodak, unclamped float chain, three rates $\times$ 24 images per codec, two rounds). Exact: round 2 has the same file bytes and symbols as round 1. $\Delta$PSNR$_2$: round-2 PSNR minus the full method's, range over rates. Random-basis and initial-encoder variants and per-rate values: Appendix Table~\ref{tab:ablation-full}.}\par\vspace{-0.1cm}
\label{tab:ablation-main}
\centering\footnotesize\setlength{\tabcolsep}{3pt}
\begin{tabular}{@{}lrrrr@{}}
\toprule
& \multicolumn{2}{c}{AEIC} & \multicolumn{2}{c}{HPCM}\\
\cmidrule(lr){2-3}\cmidrule(lr){4-5}
Variant & Exact & $\Delta$PSNR$_2$ & Exact & $\Delta$PSNR$_2$\\
\midrule
Full method & 72/72 & 0 & 72/72 & 0\\
No target search & 0/72 & $-0.39$ to $-0.59$ & 0/72 & $-0.60$ to $-0.67$\\
Wrapper only ($\hat y$) & 0/72 & $-0.29$ to $-0.52$ & 0/72 & $-0.30$ to $-0.56$\\
\bottomrule
\end{tabular}
\vspace{-0.3cm}
\end{table}

\begin{table}[!t]
\caption{\small\textbf{Generative decoding under 50 rounds of 8-bit recompression.}
FFHQ, 1,000 images, $.3718$\,bpp ($.3789$ framed).
Ours and Xu share ELIC weights and prior.
Generator-free regular decoding nearly matches ELIC's initial PSNR;
NLBP greatly improves initial FID.
At round 50, regular leads PSNR and NLBP leads FID, illustrating
the perception--distortion trade-off.
Ours re-encodes with $E'$, Xu with ELIC;
NLBP uses a code-derived seed, Xu unseeded guidance.
Exact file: matches round 1 throughout.
BPP drift: round 50 minus 1.
Bold/underlined: best/second quality per round.}
\par\vspace{-0.1cm}
\label{tab:ffhq-main}
\centering\small
\setlength{\tabcolsep}{2pt}
\renewcommand{\arraystretch}{1.0}
\setlength{\aboverulesep}{1pt}
\setlength{\belowrulesep}{1pt}
\begin{tabular*}{\linewidth}{@{\extracolsep{\fill}}lrrrrr@{}}
\toprule
Decoder & PSNR$\uparrow$ & LPIPS$\downarrow$ &
FID$_{64}\downarrow$ & \shortstack{BPP\\drift} & Exact file\\
\midrule
\multicolumn{6}{@{}l}{\textbf{Round 1}}\\
ELIC
  & \textbf{34.87} & .0748 & 37.85 & & \\
Xu et al.~\cite{xu2024}
  & 32.72 & \textbf{.0294} & \textbf{4.11} & & \\
Ours: regular
  & \underline{34.85} & .0748 & 36.29 & & \\
Ours: NLBP
  & 32.83 & \underline{.0303} & \underline{5.16} & & \\
\midrule
\multicolumn{6}{@{}l}{\textbf{Round 50}}\\
ELIC
  & 31.10 & .1285 & 59.30 & $-.070$ & 0/1000\\
Xu et al.~\cite{xu2024}
  & 24.88 & .1473 & 36.38 & $+.131$ & 0/1000\\
Ours: regular
  & \textbf{34.85} & \underline{.0748} & \underline{36.29}
  & \textbf{0} & \textbf{1000/1000}\\
Ours: NLBP
  & \underline{32.83} & \textbf{.0303} & \textbf{5.16}
  & \textbf{0} & \textbf{1000/1000}\\
\bottomrule
\end{tabular*}
\end{table}

\subsection{Recompression without retraining}
\label{sec:idem-results}
\paragraph{Coverage and exactness.}
Table~\ref{tab:adaptation-summary} summarizes all recompression protocols. Every adapted floating-point chain keeps its first file (GLC: its symbols) through all evaluated rounds, and no original codec keeps its first file. Under 8-bit storage, adapted AEIC, HPCM, ELIC, and MS-ILLM preserve their first image with zero PSNR drift through 50 rounds on Kodak (Table~\ref{tab:adaptation-summary}; Appendix Fig.~\ref{fig:codec-recompression-panel}). With the full procedure, AEIC, ELIC, and MS-ILLM also preserve every first file. HPCM preserves $137/144$ first files; the other seven change once at round 2 without changing the image. AEIC also preserves both image and file through 50 rounds in all 205 CLIC and 500 DIV2K chains, including seven DIV2K cases recovered by a fallback developed on those cases (Appendix~\ref{sec:high-resolution}).

\paragraph{First-pass cost and drift.}
Adaptation adds no initial bitrate. Table~\ref{tab:adaptation-summary} reports initial PSNR changes; HPCM's largest single-image loss is $1.80$\,dB under 8-bit storage. All six quality metrics remain unchanged through round 50 in Table~\ref{tab:recompression-main}.

\begin{figure}
\centering\includegraphics[width=0.95\linewidth, trim={0, 0.7cm, 0, 0}, clip]{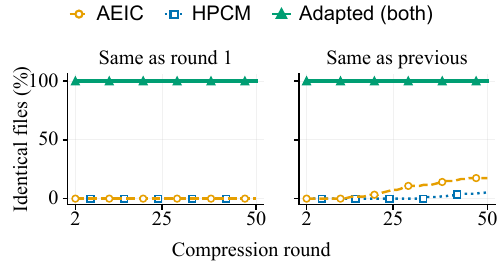}
\par\vspace{-0.2cm}\caption{\small\textbf{Preserving the transmitted bits.} Complete-file equality on Kodak, measured at every round through 50, across all 120 AEIC and 144 HPCM image--rate pairs. Both adapted curves coincide at 100\%. Some original chains later repeat a code, but none retain their initial code. Float-chain checks precede the 8-bit tests.}
\label{fig:code-equality}
\vspace{-0.2cm}
\end{figure}

\paragraph{Comparison with consistency-focused codecs.}
At an initial $.708$\,bpp, interpolated float-chain results show adapted HPCM retaining $37.03$\,dB through round 50, versus $34.89$\,dB for MSINN~\cite{msinn2025}. Published results for Li's near and exact models~\cite{rightinverse} are $34.92$ and approximately $34.50$\,dB, respectively (Table~\ref{tab:consistency-methods}).

Under 8-bit recompression, adapted HPCM preserves its first image across all tested rates (Figure~\ref{fig:idem-rd}). At round 50, it retains $33.95$\,dB versus $31.90$\,dB for MSINN, at initial rates of $.378$ and $.429$\,bpp, respectively. Li's exact model remains stable in its native float setting; its degradation under 8-bit storage concerns a different protocol. Appendix~\ref{sec:baseline-audit} provides details and further comparisons.

\begin{figure}[b]
\centering
\resizebox{\linewidth}{!}{%
\begin{tikzpicture}[
  x=1bp,y=1bp,
  font=\sffamily\fontsize{7.5}{8.5}\selectfont
]
\path[use as bounding box](0,-7)rectangle(228,124);

\foreach \xx/\sx/\rr/\plain in {
  19/250/1/35.52,
  71/778/20/31.31,
  123/1042/50/28.70,
  175/1306/100/26.29
}{
  \node at ({\xx+25},118){Round \rr};

  \begin{scope}[overlay]
    \clip(\xx,63)rectangle({\xx+50},113);
    \node[anchor=north west,inner sep=0pt]
      at ({\xx-\sx*0.1953125},{113+7.421875})
      {\includegraphics[width=308.203125bp]
        {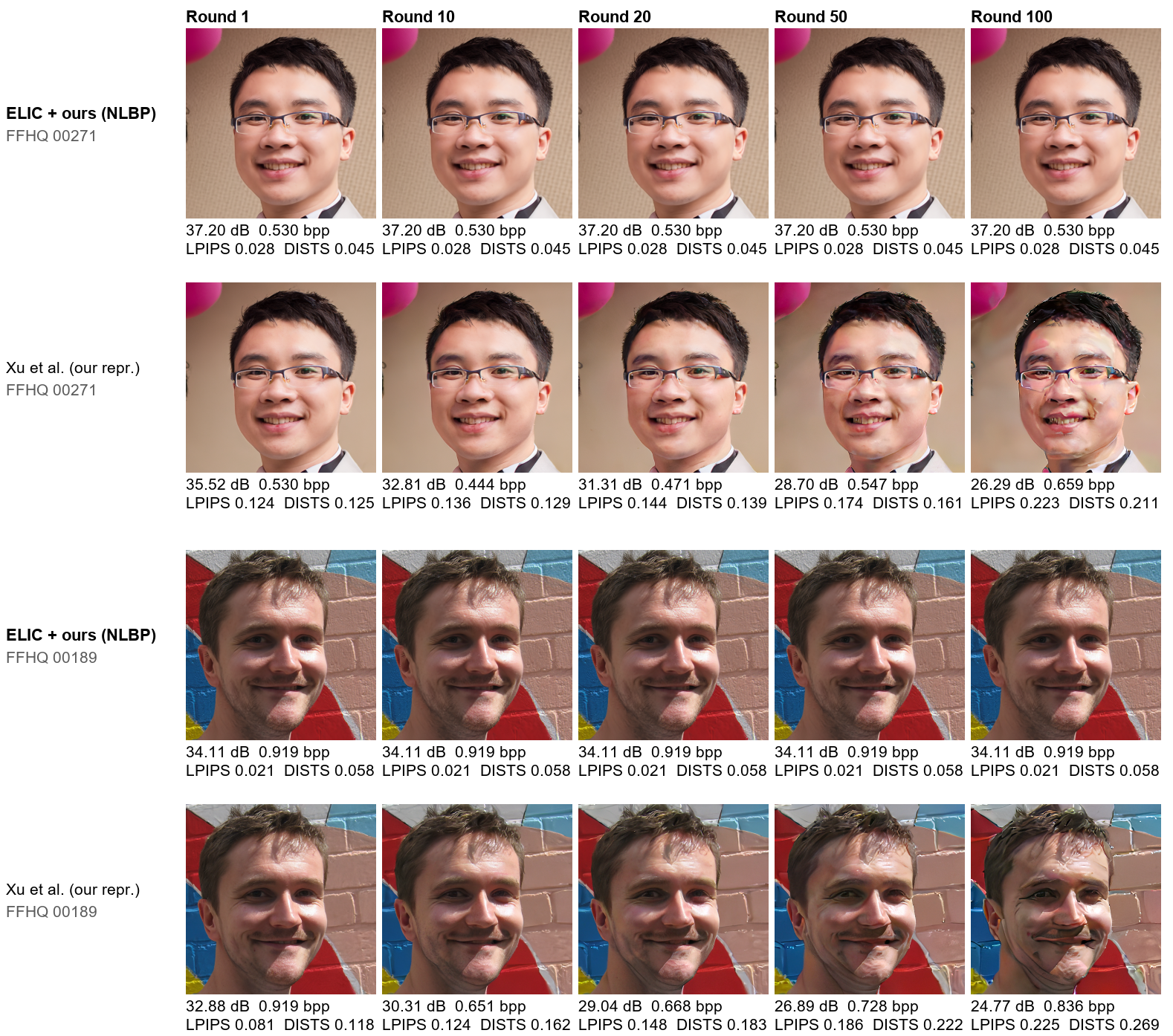}};
  \end{scope}
  \node at ({\xx+25},58){37.20};

  \begin{scope}[overlay]
    \clip(\xx,3)rectangle({\xx+50},53);
    \node[anchor=north west,inner sep=0pt]
      at ({\xx-\sx*0.1953125},{53+74.21875})
      {\includegraphics[width=308.203125bp]
        {figures/nlbp_xu_100rounds.png}};
  \end{scope}
  \node at ({\xx+25},-2){\plain};
}

\node[rotate=90,text=green!45!black] at (8,88){Ours: NLBP};
\node[rotate=90] at (8,28){Xu et al.};
\end{tikzpicture}}

\par\vspace{-0.1cm}
\caption{\small\textbf{Stable generative decoding.}
FFHQ 00271, initially $.530$\,bpp; 100 rounds, 8-bit storage.
Labels: PSNR (dB).
Selected for DISTS gain over Xu
(Appendix Fig.~\ref{fig:nlbp-xu100}).
Aggregate recompression: Appendix Fig.~\ref{fig:nlbp-recompression};
five-rate quality: Appendix Table~\ref{tab:nlbp-quality}.}
\label{fig:ffhq-main}
\end{figure}

\subsection{Decoding with a chosen generator}
\label{sec:control-results}
The receiver chooses a prior and its influence after encoding, without weight updates. Figure~\ref{fig:teaser}(c) varies starting noise at fixed transmitted representations, demonstrating perception--distortion control. Its ELIC configuration differs from the final comparison below (Appendix~\ref{sec:decoder-controls}). Figure~\ref{fig:generator-choice} decodes one HPCM bitstream with PixNerd~\cite{pixnerd2026}, Z-Image~\cite{zimage} and FLUX.2 [dev]. PixNerd restores water texture with tuned settings; the others use untuned settings.

The final ELIC/FFHQ setting was chosen on 100 tuning images and tested on 1,000 disjoint images (Appendix Table~\ref{tab:nlbp-quality}). At $.1644$\,bpp, NLBP reduces FID from $63.50$ to $7.40$ for $1.96$\,dB of PSNR. Across five rates (Appendix Table~\ref{tab:nlbp-quality}), final 8-bit code checks pass on $1{,}000/1{,}000$ images at the first four and $990/1{,}000$ at the highest; failures remain included.

\paragraph{Matched comparison.}
With matched ELIC weights and prior, NLBP has higher PSNR and Xu~\cite{xu2024} lower FID across five rates. Over 50 rounds of 8-bit recompression on 1,000 FFHQ images, NLBP preserves every first file, retaining $32.83$\,dB and FID$_{64}$ $5.16$; Xu falls to $24.88$\,dB with FID$_{64}$ $36.38$ (Table~\ref{tab:ffhq-main}; Appendix Fig.~\ref{fig:nlbp-recompression}).

\subsection{Ablations and decoding cost}
\label{sec:cost}
Table~\ref{tab:ablation-main} isolates components. The wrapper alone and main-cell projection without the hyper constraint each give $0/72$ exact pairs per codec after one recompression, versus $72/72$ for full search. A random orthonormal basis preserves most codes ($72/72$ AEIC, $70/72$ HPCM) but harms quality (HPCM round 2: $-12.1$ to $-16.5$\,dB): PCA coordinates matter. Initial $E'$ encoding stays exact but changes the file and rate (HPCM: $+3$ to $+16\%$; Appendix Table~\ref{tab:ablation-full}). Further 8-bit repair ablations evaluate hard-case recovery and target-update efficiency (Appendix, Table~\ref{tab:ablation-8bit}).

The cost is at the receiver and arises already in regular decoding, before any generator; initial encoding is unchanged. Median adapted decoding of a Kodak image takes $2.6$--$3.9$\,s (AEIC) and $1.2$--$6.5$\,s (HPCM) in floating point, against about $0.1$\,s originally, and $5.3$--$8.0$\,s and $5.0$--$18.7$\,s under 8-bit storage on slower nodes, at most 253\,s over 7,200 HPCM decodes. NLBP adds its sampler's time. Timer scopes differ across protocols (Table~\ref{tab:runtime-placeholder}, Appendix~\ref{app:cost-details}).

\section{Limitations}
\vspace{-0.1cm}
\label{sec:limitations}
Decoder-side search and 8-bit repair take seconds to tens of seconds per Kodak image and may fail. Initial encoding uses $E$, while recompression uses $E'$, requiring the application to track reconstructed inputs.
Using $E'$ throughout removes this requirement but can increase the initial bitrate (up to $2.7\times$ for AEIC at $c=10$; Appendix Table~\ref{tab:png-first-encoder}). Preservation is certified only for accepted targets on the tested platform; 8-bit search success is not guaranteed, and image equality need not imply first-file equality. High-resolution recompression requires the padded canvas and retained colour metadata. NLBP may introduce texture artifacts and yields higher first-pass FID than guidance~\cite{xu2024} in our comparison. Kodak FID estimates are limited by the small dataset.

{\small \paragraph{Acknowledgements.} A.S. is a Chaya Fellow and acknowledges support from the Chaya Career Advancement Chair and the Alon Fellowship of the Israeli Council for Higher Education. This research was supported by the Israel Science Foundation, Grant No. 1039/26. This research was made possible through a GPU compute resource grant funded by the Association of University Heads, the Council for Higher Education and the AI Research Compute Center.

We sincerely thank Ron Libman and Yamit Ehrlich \mbox{(Technion)} for helpful discussions.
\par}

\clearpage
{\small\bibliographystyle{ieeenat_fullname}\bibliography{references}}
\ifincludeappendix
\clearpage
\appendix
\renewcommand{\topfraction}{0.95}\renewcommand{\dbltopfraction}{0.95}
\renewcommand{\textfraction}{0.04}\renewcommand{\floatpagefraction}{0.85}
\setcounter{topnumber}{3}\setcounter{dbltopnumber}{3}\setcounter{totalnumber}{5}

\section{Implementation and Evaluation Protocols}
\label{sec:full-results}
\subsection{Codec-Specific Implementations and Settings}
\label{sec:implementation-settings}
All codec weights stay frozen, so pretrained checkpoints are adapted without retraining or fine-tuning; the only fitted component is a patch-PCA basis. AEIC uses a $32\times32$ basis fitted on the 585 CLIC~2020 Professional training images, and all other codecs share a $16\times16$ basis fitted on 1,000 Vimeo90K frames; neither set overlaps an evaluation set. The weak coordinates are the $k$ lowest-variance directions, with $k$ the codec's latent channel count, and each basis is reused unchanged across rates, protocols, and generative decoders, so two fixed bases cover all eight codecs. Table~\ref{tab:png-settings} lists all per-codec settings, and Appendix~\ref{sec:technical-details} gives the search and repair mechanics. GLC~\cite{glc2024}, which has no entropy coder, is searched over vector-quantization codewords and evaluated in the float chain only.

Under 8-bit storage, decoding follows a fixed escalation ladder that is part of the method: preset correction, a doubled correction budget, a wider hyper guard, and an image-space box search, each invoked only after the earlier stages fail (Appendix~\ref{sec:png-details}). The stages change only the search; all use the same acceptance test, namely that re-encoding the stored 8-bit image reproduces the received symbols. Refinement uses learning rate $0.003$ and penalty weight $100$, and HPCM clipped-float results are limited to $\lambda\le.013$, since higher rates require $c=30$.

\textbf{No setting is chosen per image or per chain}, and the decoder never accesses the source image. All settings were developed on Kodak (AEIC, HPCM, ELIC) and DIV2K (AEIC), where 891 of the 908 8-bit chains (98.1\%) were already exact with the stage-0 preset; the escalation stages were designed on ten of the remaining failures (seven DIV2K, three ELIC); HPCM, implemented before the ladder, handles its seven by outer stabilization, which no other codec uses. The settings and the ladder were then frozen: on every other experiment (MS-ILLM, Hyperprior, Mean-scale, TCM, AEIC on CLIC, and the FFHQ 50-round chains at the fourth rate), 1,801 of the 1,805 8-bit chains (99.8\%) are exact with the stage-0 preset alone. The 65 ELIC chains of the $256^2$ comparison also pass with the preset alone.

\begin{table*}[t]
\centering\small
\caption{\small\textbf{Implementation settings.} Training-free wrapper, target search, and 8-bit repair for each codec. Protocols evaluated in the paper: F float, C clipped float, 8 8-bit storage. PCA bases are fitted on CLIC~2020 Professional ($32\times32$ patches) or Vimeo90K ($16\times16$). The ELIC column also applies to the $256^2$ and FFHQ experiments, which use Xu et al.'s ELIC weights. Clipped-float runs (AEIC, HPCM) add at most 200 range-repair iterations with an inward guard of $10^{-4}$ of the range. Budgets count search steps, repair updates, or evaluated candidates, not recompression rounds. $^{a}$Fraction of the quantization step. $^{b}$Squared-distance gap to the eight nearest competing codewords. $^{c}$DIV2K only (2K canvas). $^{d}$Hyperprior and Mean-scale: CompressAI zoo (MSE); TCM: official release. $^{e}$Latent channels: 192 or 320 by quality for Hyperprior and Mean-scale, 320 for TCM.}
\label{tab:png-settings}
\setlength{\tabcolsep}{4pt}
\begin{tabular}{@{}lcccccc@{}}
\toprule
Setting & AEIC & HPCM & ELIC & MS-ILLM & \shortstack{Hyperprior,\\Mean-scale, TCM} & GLC\\
\midrule
Weights & official & Large, MSE & re-implementation & official & CompressAI$^{d}$ & official\\
Protocols & F, C, 8 & F, C, 8 & F, 8 & 8 & 8 & F\\
Image range & $[-1,1]$ & $[0,1]$ & $[0,1]$ & $[0,1]$ & $[0,1]$ & $[-1,1]$\\
\addlinespace[2pt]
PCA basis, patch width & CLIC, 32 & Vimeo, 16 & Vimeo, 16 & Vimeo, 16 & Vimeo, 16 & Vimeo, 16\\
Weak coefficients $k$ & 320 & 320 & 320 & 220 & 192--320$^{e}$ & 256\\
Gain $c$: F / C, 8 & 1 / 10 & 10 / 10 & 10 / 10 & -- / 10 & -- / 10 & 10 / --\\
\addlinespace[2pt]
Search start & projected & projected & projected & projected & projected & received\\
Steps / learning rate & 400 / .01 & 4,000 / .003 & 400 / .01 & 400 / .01 & 400 / .01 & 400 / .002\\
Main margin $m_y$: F / 8 & .02 / .12 & .02 / .06 & .02 / .06 & -- / .06 & -- / .06 & .02$^{a}$ / --\\
Hyper margin $m_z$ & .02 & .002 & .02 & .02 & .02 & $10^{-4\,b}$\\
Refinement steps & 1,000 & 200 & 200 & 200 & 200 & --\\
\addlinespace[2pt]
8-bit correction budget & 100 (200$^{c}$) & 30 & 30 & 30 & 30 & --\\
Lattice trials per call & -- & 9,056 & 9,056 & 9,056 & 9,056 & --\\
Per-symbol hyper repair & -- & yes & yes & yes & yes & --\\
Escalation stages & 1--2$^{c}$ & -- & 0--3 & -- & 0--3 & --\\
Outer stabilization & -- & 4 states & -- & -- & -- & --\\
\bottomrule
\end{tabular}
\end{table*}

\subsection{Recompression Protocols}
\label{sec:recompression-protocols}
Three representations are passed between compression rounds:
\begin{itemize}
\setlength{\itemsep}{0pt}
\item \textbf{8-bit:} the reconstruction is clipped, rounded to 8-bit RGB, and converted back with the next encoder's normalization.
\item \textbf{Clipped float:} the reconstruction is clipped to the valid image range without pixel quantization.
\item \textbf{Float:} the adapted decoder's output is passed to the next encoder without an additional clamp; base codecs keep their native decoding operations.
\end{itemize}
For coordinates in $[0,1]$, the 8-bit conversion is
\begin{equation}
S_8(x)=\frac{\operatorname{rint}(255\,\operatorname{clip}(x,0,1))}{255},
\label{eq:png-format}
\end{equation}
with round-to-nearest, ties-to-even. For AEIC, the conversion includes the mapping between $[-1,1]$ and $[0,1]$. Lossless PNG storage preserves the integer pixels. Consistency does not transfer across representations: over 50 rounds, the exact model of Li et al.\ has zero PSNR drift in its native float chain but drifts by $-15.56$\,dB under clipping and $-20.91$\,dB under 8-bit storage, whereas every tested chain of our method keeps its first reconstruction in all three protocols.

All 8-bit chains except adapted AEIC on Kodak are file chains: each round writes the compressed file and the reconstructed PNG to disk, and the next round decodes and encodes exactly these files. Exactness in these chains therefore covers entropy coding, file serialization, and pixel quantization, not only tensor-level consistency. This covers HPCM, ELIC (Kodak, $256^2$, FFHQ), MS-ILLM, Hyperprior, Mean-scale, TCM, AEIC on DIV2K and CLIC, and the Xu et al. chains, which in each later round decode ELIC's encoding of their previous output. Adapted AEIC on Kodak passes the 8-bit image in memory instead of through a file; on 12 chains over 10 rounds, this gives the same files and images as writing and reloading the PNG. Earlier HPCM in-memory measurements are identified separately where used.

The first encode of every chain uses the base encoder; later encodes of our chains use the wrapped encoder $E'$, which the runner selects by round. The first file is therefore the base codec's own file: the initial rate is unchanged, and an already-encoded image can enter a chain without its source. Applying $D'$ to the identity $E'(D'(E'(x)))=E'(x)$ shows that the image-to-image operator $D'\circ E'$ is idempotent on every input whose decode passes the check; under 8-bit storage the operator is $D_8\circ\Sigma\circ E'$. Our chains apply this one operator from round 2 on, and because it returns every accepted reconstruction unchanged, each later round returns the first reconstruction. Selecting $E'$ requires identifying decoded reconstructions. Two simple mechanisms do so without changing the evaluated chains: a one-bit flag stored with the decoded image, or, without side information, a fixed-point self-detection test that encodes the input with $E'$, decodes, and selects $E'$ only if the decode returns the input. The test's outcome on certified reconstructions is measured in all 146,902 recompressions of the summary table, since its decode is the computation of the next round of our chains; a failing input is encoded with $E$ as a new source, and a passing source image is reproduced exactly by $E'$. The test requires the decoder and its configuration at the encoder; efficient detection is left to future work. Using $E$ for the first encode is a practical choice, and idempotency does not depend on it. Re-encoding a reconstruction with the base encoder $E$ is outside this argument and was not evaluated. Using $E'$ from the first encode removes the distinction and also stays exact, at a cost that depends on the gain: in the float chain ($c=1$), AEIC's first rate changes by $-0.25$ to $+0.69\%$; under 8-bit storage ($c=10$), all 120 AEIC chains remain file-exact through 50 rounds, but the initial rate rises by factors of $1.26$--$2.74$ and the rate-wise mean PSNR changes by $-0.07$ to $-0.20$\,dB. The ablation of a wrapped first encode is reported in Appendix~\ref{sec:first-encoder-ablation}.

\subsection{Datasets and High-Resolution Evaluation}
\label{sec:datasets}
Kodak evaluation uses all 24 images at their original resolution. The evaluated rates and measured recompression horizons are listed with each result in Table~\ref{tab:grand-all}.

The $256\times256$ comparison with Xu~\cite{xu2024} feeds the same preprocessed inputs to Xu, plain ELIC, and ELIC with our adaptation: all 24 Kodak images and the 41 CLIC images of the high-resolution evaluation, each center-cropped to a square and resized with PIL bicubic interpolation.

High-resolution evaluation uses 100 DIV2K validation images and 41 CLIC Pro validation images. The AEIC protocol retains the padded canvas and the six color-statistic integers from the first code; they are transmitted as side information in every round and counted in the payload, and the next encoder receives the retained canvas.

FFHQ experiments use $256\times256$ images: generative-decoding settings are selected on 100 tuning images (01000--01099), frozen, and evaluated on 1,000 disjoint test images (00000--00999). Prior-specific settings and the selection of qualitative examples are described in Appendix~\ref{sec:nlbp-full}.

\subsection{Metrics, Rate Accounting, and Determinism}
\label{sec:metrics}
We report PSNR, MS-SSIM, LPIPS, DISTS, FID, and KID, computed on the stored pixels for 8-bit experiments and on clipped reconstructions for float experiments. PSNR is averaged over images. FID and KID use nonoverlapping $64\times64$ patches: 2,304 for Kodak at native resolution and 16,000 for FFHQ, where FID is computed with Xu et al.'s code. For the $256^2$ comparison, the same patches give small-sample estimates (16 per image). High-resolution evaluation uses $256\times256$ patches. For GLC's highest rate, paired quality comparisons use the same 23 completed images and 2,208 patches; equality counts keep their own denominator. Distributional scores are compared only within the same dataset, image subset, and scoring protocol. KID is unscaled unless specified.

\textbf{Image-exact} means that every evaluated reconstruction equals the first reconstruction in the chain's representation. \textbf{File-exact} means that every later compressed file equals the initial compressed file, including all streams and headers; this refers to codec files (the compressed bitstreams), not PNG container bytes. File equality is checked at every round. Image equality is either measured (pixel or tensor comparison) or inferred from identical files and deterministic decoding; the two, like symbol and coded-stream checks, are reported separately. Under 8-bit storage, all 1,778 adapted chains in the summary table are image-exact and 1,771 are file-exact; the other seven, all HPCM, change file once at round 2. The base codecs are exact in none of these chains, and none of the 281 chains of Li et al.\ (outside its native float chain), MSINN, or Xu et al.\ is image- or file-exact.

The first-round quality change is
\begin{equation}
\Delta\mathrm{PSNR}_1=\mathrm{PSNR}_{\mathrm{ours},1}-\mathrm{PSNR}_{\mathrm{base},1}.
\end{equation}
Recompression drift compares the last and first rounds of the same method and protocol. Rounded metric equality is not an exactness test: exactness is established from file, symbol, pixel, or tensor comparisons, never from a rounded score.

Payload rate includes all entropy-coded streams and transmitted reconstruction side information, excluding container framing, and is normalized by the original image dimensions. GLC has no entropy coder, so its rate is estimated and its consistency checks are at the symbol level. Published reference values keep their stated rate conventions.

We verified exactness only on the hardware and software used for each run; other platforms were not tested. Regular decoding uses no randomness; NLBP derives its seed from the received symbols, and the Xu reproduction uses an unseeded sampler. Hardware, timing scope, and computational cost are reported in Appendix~\ref{app:cost-details}.

\FloatBarrier
\section{Technical Details and Guarantees}
\label{sec:technical-details}
\label{sec:background-details}
\subsection{Wrapper Construction, Completion, and Gain}
\label{sec:spnn-prelim}
\label{sec:linear-bp}
\label{sec:nlbp-prelim}
\label{sec:completions}
\label{sec:gain-fidelity}
\label{sec:gain-calibration}
For orthonormal image coordinates $a(x)$ and $b(x)$, the wrapped encoder and exact write are
\begin{align}
E'(x)&=f(b(x))+c\,a(x),\nonumber\\
W(v,b)&=J\!\left(\frac{v-f(b)}c,b\right).
\end{align}
Substitution gives $E'(W(v,b))=v$ for every $c>0$ in real arithmetic. Retaining $b(x)$ therefore makes $G(x)=(E'(x),b(x))$ a bijective completion. Its zero-complement right inverse is $W(v,0)$.

More generally, a completion
\begin{equation}
G(x)=(g(x),q(x))\label{eq:prelim-completion}
\end{equation}
defines
\begin{equation}
g^\dagger(v)=G^{-1}(v,0).\label{eq:prelim-pinv}
\end{equation}
It satisfies $gg^\dagger=\mathrm{Id}$, and hence both reflexive identities $gg^\dagger g=g$ and $g^\dagger gg^\dagger=g^\dagger$. The inverse depends on the completion~\cite{spnn}. Our construction instantiates this structure with frozen pretrained networks~\cite{dejavae}. That wrapper covers continuous transforms; quantized symbols, hyper side information, entropy-coded files, and 8-bit storage are addressed in the following subsections.

\paragraph{Back-projection.}
For feasible linear measurements, back-projection gives
\begin{equation}
u^+=u+A^\dagger(v-Au),\label{eq:linear-bp}
\end{equation}
the closest consistent image in Euclidean distance. For a nonlinear completion and a right inverse $R$, the corresponding update is
\begin{equation}
u^+=G^{-1}\!\left(G(u)-G(R(g(u)))+G(R(v))\right).
\label{eq:generic-nlbp}
\end{equation}

With the completion-induced inverse, this simplifies to
\begin{equation}
u^+=G^{-1}(v,q(u)),\label{eq:prelim-coordinate-bp}
\end{equation}
the measurement coordinates are replaced and the complementary coordinates are retained. The closest-point interpretation applies in completion coordinates, not generally in pixel space.

\paragraph{Completion choice.}
For our wrapper, choosing $R_r(v)=W(v,r(v))$ gives
\begin{equation}
u^+=W\!\left(y^\star,b(u)-r(E'(u))+r(y^\star)\right).
\end{equation}
Both endpoints must use the same reference mapping $r$. Configurations using different quantization conventions at the two endpoints still admit a final exact write, but need not implement this single-right-inverse formula.

\paragraph{Gain and fidelity.}
Let $\Delta(x)=E(x)-E(P_1x)$, $a_D(v)=a(D(v))$, and $e_D(v)=v-E(P_1D(v))$. Orthogonality gives
\begin{align}
E'_c(x)-E(x)&=c\,a(x)-\Delta(x),\label{eq:gain-encoder}\\
\|D'_c(v)-D(v)\|_2&=\|e_D(v)/c-a_D(v)\|_2.\label{eq:gain-decoder}
\end{align}
Increasing $c$ amplifies the encoder's weak-coordinate contribution and reduces the decoder residual term. The total errors need not change monotonically because the terms can cancel.

\paragraph{Gain calibration.}
On fixed calibration distributions, let $\sigma_x,\tau_x,\sigma_D,\rho_D$ denote the RMS norms of $a,\Delta,a_D,e_D$. Minkowski's inequality gives the sufficient interval
\begin{equation}
\frac{\rho_D}{\varepsilon_D-\sigma_D}\le c\le\frac{\varepsilon_E-\tau_x}{\sigma_x},
\end{equation}
when the required quantities are positive and the interval is nonempty. An empty interval means that these bounds cannot certify the requested tolerances.

A weighted squared-error objective has the form
\begin{equation}
\mathcal J(c)=Ac^2-2Bc+C/c^2-2K/c+\mathrm{const},
\end{equation}
where $A=\lambda_E\mathbb E\|a\|^2$, $B=\lambda_E\mathbb E\langle a,\Delta\rangle$, $C=\lambda_D\mathbb E\|e_D\|^2$, and $K=\lambda_D\mathbb E\langle a_D,e_D\rangle$. For $A,C>0$, candidate minima are positive roots of
\begin{equation}
Ac^4-Bc^3+Kc-C=0,
\end{equation}
together with any constrained interval endpoints. These are continuous calibration rules; the experiments use the fixed gains in Appendix~\ref{sec:implementation-settings}.

\subsection{Target Search and Quantization Constraints}
\label{sec:target-search}
\label{app:main-cell}
\label{app:toy-example}
\label{app:feasible-nonempty}
A quantization cell contains all continuous values producing the same symbol. For coordinatewise unit-step rounding, fixing the received symbols fixes the quantization offsets and reconstructed context. The main cell is therefore a box centred at $\hat y$, with interior
\begin{equation}
\operatorname{int}\mathcal C_y(s)=\{t:\|t-\hat y\|_\infty<1/2\}.
\label{eq:main-cell-interior}
\end{equation}

Boundary membership follows the codec's rounding convention. The hyper constraint is imposed after the nonlinear $h_a$, so $\mathcal C_z(s)$ need not be a box or convex. Vector-quantized streams instead use codeword regions.

\paragraph{Why writing $\hat y$ is insufficient.}
Writing $\hat y$ alone may change the hyper symbols. For example, let $Q_y(y)=\operatorname{round}(y)$ and $h_a(y)=0.7y$ in a scale-only hyperprior~\cite{balle2018}. The latent $y=2.4$ produces main and hyper symbols equal to 2, but writing $\hat y=2$ produces hyper symbol 1. Targets in $[2.15,2.5)$ preserve both.

\paragraph{Continuous feasibility.}
The continuous feasible set is nonempty because the sender's original latent produced the received symbols. This witness need not belong to the margin-constrained set and is unavailable to the receiver. The receiver therefore does not attempt to recover the sender's latent: any target in the feasible set reproduces the received symbols.

The search projects $E'(x_{\mathrm{raw}})$ into the shrunk main-cell box. Each iterate is projected back into this box, so the objective acts only on the hyper constraint and both symbol constraints are handled together. For scalar hyper cells, it minimizes $\sum_j(v_j(t)+0.01\,\mathbf1_{v_j(t)>0})^2$, where
\begin{equation}
v_j(t)=\left[|h_a(t)_j-\hat z_j|-(1/2-m_z)\right]_+.
\end{equation}
A target is accepted only when the margin conditions and the codec's actual quantizers agree with the received record.

Refinement moves the feasible target toward the projected start, using 12 bisection steps followed by bounded optimization and retaining the nearest feasible iterate found. It uses an additional hyper-cell guard of $0.01$. If the projected search fails, the decoder restarts at $\hat y$. If neither search succeeds, it writes $\hat y$, which preserves the received record only when that target is itself feasible.

The search uses received symbols and frozen networks. It does not require the source image or guarantee a globally closest target.

\subsection{Clipped-Float and 8-bit Repair}
\label{sec:png-details}
\label{sec:clamp-repair}
\label{app:escalation}
\paragraph{Clipped float.}
Starting from $x_0=W(y^\star,b)$, range repair alternates clipping and writing:
\begin{equation}
x_{k+1}=W\!\left(y^\star,b(\Pi(x_k))\right).\label{eq:range}
\end{equation}
Each write restores the continuous target in real arithmetic. If an iterate lies within the valid range, the chain's clipping step leaves it unchanged. Otherwise, the bounded procedure may terminate without a consistent output; the final operational check determines success. With the reported settings (five AEIC rates and four HPCM rates), all 216 clipped-float chains keep their initial file through 50 rounds.

\paragraph{8-bit target correction.}
Each trial is formatted before re-encoding, so each update responds to the latent of the stored 8-bit image rather than of the continuous output. For $v_k=E'(x_k)$, define
\begin{align}
r_y(v)&=\operatorname{sign}(v-\hat y)\odot[|v-\hat y|-(1/2-\gamma_y)]_+,\nonumber\\
L_z(v)&=\sum_j[|h_a(v)_j-\hat z_j|-(1/2-\gamma_z)]_+^2.\nonumber
\end{align}
The target update is
\begin{equation}
t_{k+1}=t_k-r_y(v_k)-\frac{L_z(v_k)}{\|\nabla L_z(v_k)\|_2^2}\nabla L_z(v_k).
\label{eq:png-target-update}
\end{equation}
The last term is omitted when its loss or gradient norm is zero. The initial correction guards are $\gamma_y=\gamma_z=0.02$. Gradients are evaluated at the realized latent; the update does not differentiate through rounding. On eight ordinary three-round chains, omitting this update retains exactness but takes $1.5$--$2.4\times$ longer.

\paragraph{Lattice and per-symbol repair.}
When enabled, lattice search ranks pixel-channel changes using the gradient of
\begin{align}
M(x)={}&\sum_j[|h_a(E'(x))_j-\hat z_j|-(1/2-\eta_z)]_+^2\nonumber\\
&+\sum_i[|E'(x)_i-\hat y_i|-(1/2-\eta_y)]_+^2,\label{eq:png-lattice-merit}
\end{align}
with $\eta_z=0.003$ and $\eta_y=0.005$. It tests one-grey-level moves in decreasing groups, preferring fewer hyper mismatches and then lower merit. Additional scans and per-symbol repair operate while hyper mismatches remain. Per-symbol repair considers up to 16 wrong hyper symbols per round, with limits of 12 rounds and 400 trials, accepting changes that reduce hyper mismatches while keeping main coordinates inside their cells.

\paragraph{Escalation.}
The decoder invokes additional stages only after a failed acceptance check. The implementation-specific budgets are listed in Appendix~\ref{sec:implementation-settings}. ELIC, Hyperprior, Mean-scale, and TCM can increase the correction budget, widen the hyper guard, and apply projected image-space box search. The AEIC DIV2K implementation starts at the larger budget and retains the better candidate if its wider-guard retry also fails. On the development sets, these stages recover the three ELIC and seven DIV2K chains that the preset does not pass. Within a stage, the decoder keeps the round with the fewest mismatched symbols; if every stage fails, the returned best-effort output is not certified as exact.

DIV2K, HPCM, and ELIC were used during development of the repair procedure. Preset and full-method results are distinguished in Appendix~\ref{sec:serialization-results}.

\paragraph{Decoder-side stabilization.}
HPCM can advance to another symbol record when repair of the received record fails:
\begin{equation}
s_{i+1}=\Sigma(E'(D_8(s_i))).\label{eq:png-stabilize}
\end{equation}
The HPCM decoder uses at most four states and one candidate variant per state. It returns a candidate when re-encoding reproduces that state's symbols. If all states fail, the returned image remains uncertified. This procedure can stabilize the image while changing the initial compressed file. In the HPCM experiments, this occurs in seven of 144 chains: each file changes once, at round 2, and the first image is retained.

\subsection{Guarantees and Limitations}
\label{sec:finite}
\label{sec:png-nonexistence}
A successful post-format check certifies that the stored reconstruction reproduces the tested symbol record. With deterministic decoding, receiving that record again returns the same image. Equal bytes additionally require deterministic entropy coding and unchanged metadata.

For stabilization, suppose the decoder accepts $x^\star$ at state $s_i$. When $s_i$ is received in the next generation, the same deterministic candidate test returns $x^\star$ immediately. This establishes an image fixed point, even when $s_i\ne s_0$.

Formatting perturbs both paths of the wrapped encoder. For $v=W(t,b)$ and $\delta=S_8(v)-v$,
\begin{equation}
E'(v+\delta)-t=cU_0\delta+f(b(v+\delta))-f(b(v)).\label{eq:png-perturbation}
\end{equation}
If $v$ is in range, rounding changes each coordinate by at most $h/2$, where $h$ is the pixel-grid spacing. With cell margins $\rho_y,\rho_z$ and local Lipschitz bounds $K_y,K_z$, the conditions
\begin{equation}
K_y h/2<\rho_y,\qquad K_z h/2<\rho_z\label{eq:png-sufficient}
\end{equation}
suffice to preserve the symbols in real arithmetic. These bounds are conditional and are not established uniformly for the evaluated networks. The implementation therefore checks the formatted output directly.

\paragraph{Continuous feasibility without an 8-bit preimage.}
Consider $E(x)=x$, $h_a(y)=0$, unit-step rounding, and the wrapper $E'(x)=510x$. The original encoder maps $x=1$ to main symbol 1, and the wrapped encoder has the continuous preimage $1/510$. However, every 8-bit input $k/255$ produces main symbol $2k$, never 1. Thus continuous feasibility does not ensure an 8-bit preimage. A continuous right inverse therefore does not by itself give 8-bit idempotence, which motivates the repair stage and its stored-image check.

For a later code produced by $E'$ from an 8-bit image, that image is a witness of existence. Decoder-side stabilization relies on this property: each advanced record is encoded from an 8-bit image and therefore has an 8-bit preimage. The witness does not establish proximity to the base reconstruction or success of a bounded search. Finite precision, clipping, and platform changes must therefore be covered by the operational checks.

\paragraph{Existence and measured success.}
Under deterministic decoding, an accepted output is certified on the stored file and image and makes the chain a fixed point; what is not guaranteed is that a solution exists or is found. In the float chain the received record has a continuous preimage, but the search uses the margin-constrained subset, which may be empty; under 8-bit storage the first record may have no 8-bit preimage. Every later record, however, is encoded by $E'$ from the previous stored image, which re-encodes to it by construction, so round 1 is the only round without a known exact solution. Because the hyper constraint passes through a nonlinear network, the search carries no general guarantee, and we measure its success on data not used for development. With all settings frozen, the stage-0 preset passes in 1,801 of the 1,805 8-bit chains run afterwards (99.8\%), and with escalation all 1,805 keep their first image; every 8-bit chain of the summary table also keeps its first image (1,778/1,778). Rounds are not independent trials, since later rounds repeat a deterministic decode, and chains that share an image need not be independent either. We therefore count images: an image fails if any of its chains fails. None of the 1,065 distinct images of the 1,805 chains (24 Kodak, 41 CLIC, 1,000 FFHQ) fails, a one-sided 95\% bound of 0.28\% on the per-image failure rate for the tested codecs and rates, assuming independent images. The bound is dominated by the 1,000 FFHQ images, which use one codec and rate; the 24 Kodak images, which carry 600 of the chains, give 11.7\%. These are recompression chains; the first-output code checks of NLBP decoding on FFHQ, of which 10 of 1,000 fail at the highest rate, are a separate test reported with the FFHQ evaluation. In HPCM, failures occur only where existence is not guaranteed: all seven failed first attempts among the 7,200 8-bit decodes are round-1 decodes, and all 7,056 later decodes pass at the first attempt. A known solution does not by itself make the bounded search succeed: chains that need escalation repeat their stage-0 failure in every round. In the float chain, the full search finds a margin-feasible target for all 144 ablation images, and all 1,004 float chains keep their first file or symbols.

Forced failures measure what a failed search costs. On 24 Kodak images over 50 rounds, an ablation returns a failing candidate whenever one exists: either the candidate closest to acceptance (the one with the fewest mismatched symbols; in clipped float, the last range-repair iterate before the accepted one) or the unrepaired output ($W(\hat y,b)$ without search in float, which the decoder also writes when no target is accepted; $W(y^\star,b)$ before range repair or 8-bit correction otherwise). Table~\ref{tab:forced-failures} reports the results. A failed search is rare: the frozen 8-bit settings fail at stage 0 in 4 of 1,805 chains, and escalation recovers all four. When it does fail, the closest candidate limits the loss. Under clipping and 8-bit storage, the unrepaired output never passes, so every decode of both arms is forced, and the closest candidate is much better: it keeps the first image in all 24 HPCM clipped-float chains, drifts less on all 24 images of both 8-bit codecs, and costs 0.02--0.03\,dB per forced round under 8-bit storage. Even with every decode forced, HPCM drifts less than the trained consistency codecs under the same protocols: under 8-bit storage it loses 1.44\,dB at 0.378\,bpp, against 2.10 and 1.97\,dB for MSINN at 0.294 and 0.429\,bpp and 20.91\,dB for Li's exact model at 0.399\,bpp; in clipped float it loses none, against 15.56\,dB for Li's model. In the float chain, where every received record has a continuous preimage and the written output is not formatted, the closest candidate is not necessarily better; it drifts less on only 9 of 24 AEIC images. Here $\hat y$ itself passes the check after a median of six rounds, so only 255 of the unrepaired arm's 1,200 decodes are forced, whereas the search visits failing iterates in every decode and all 1,200 closest-candidate decodes are forced; per forced decode, the closest candidate loses 0.03\,dB and the unrepaired output 0.13\,dB. The full method keeps the first image in all four settings, and no forced chain enters the summary table.

\section{Extended Recompression Results}
\label{sec:serialization-results}
\subsection{Float Results Across Codecs and Rates}
\label{sec:float-results}
Every chain reported in this section runs 50 rounds, one compression followed by 49 recompressions. Across all 2,998 adapted image--rate chains of the summary table (1,778 under 8-bit storage, 216 clipped-float, and 1,004 float; 146,902 recompressions), PSNR drift is zero and every chain keeps its first reconstruction (100\%). The first file (GLC: the symbols) is kept in 2,991 chains (99.8\%; 8-bit storage: 1,771 of 1,778, 99.6\%); in the remaining seven 8-bit HPCM chains it changes once, at round 2. No base decoder keeps its first file in any chain (0\%). Of the 1,778 8-bit chains, 908 belong to the development sets (Kodak AEIC, HPCM, and ELIC; DIV2K AEIC), including all seven chains whose file changes; the other 870 ran with frozen settings, and all keep their first file and image. Table~\ref{tab:bd-rate} expresses the first-pass cost as BD-rate against the base codec: at most 1.6\% except Hyperprior and Mean-scale (5.4\% and 4.3\%). The float experiments evaluate AEIC, HPCM, ELIC, and GLC through 50 recompression rounds. Adapted AEIC, HPCM, and ELIC preserve the initial files in all 120, 144, and 144 image--rate chains, respectively; GLC preserves its symbols in all 96 chains and is evaluated with estimated rates.

The largest rate-wise mean first-round PSNR changes are $-0.0024$\,dB (AEIC), $-0.138$\,dB (HPCM), $-0.026$\,dB (ELIC), and $+0.007$\,dB (GLC). These means describe the one-time adaptation cost; the PSNR drift after round 1 is zero in every adapted float chain.

\begin{figure*}[t]
\centering\includegraphics[width=\textwidth]{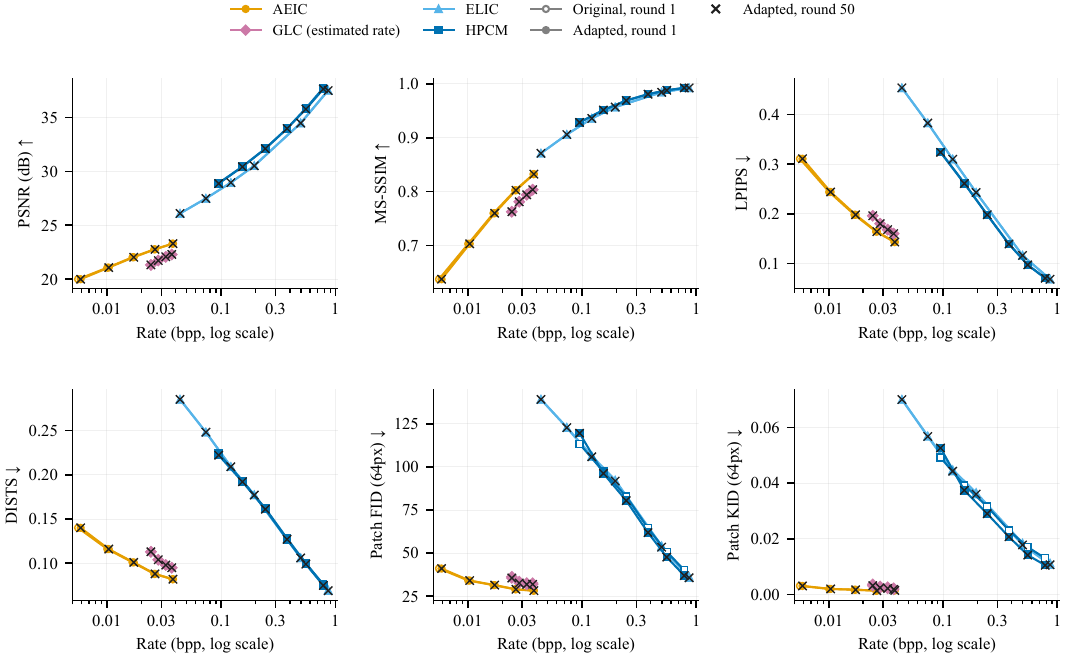}
\par\vspace{-0.1cm}\caption{\small\textbf{All six metrics at every evaluated Kodak rate.} Original and adapted first reconstructions, with the adapted round-50 reconstruction shown for all four codecs. Overlapping markers indicate unchanged scores. Operational rates count coded payload and transmitted side information, excluding framing; GLC bitrate is estimated. At GLC q3, paired quality scores use the same 23 images that complete the original chain. Published reference points, where shown, retain their labelled source convention.}
\label{fig:allmetrics}
\end{figure*}

\begin{figure*}[t]
\centering\includegraphics[width=\textwidth]{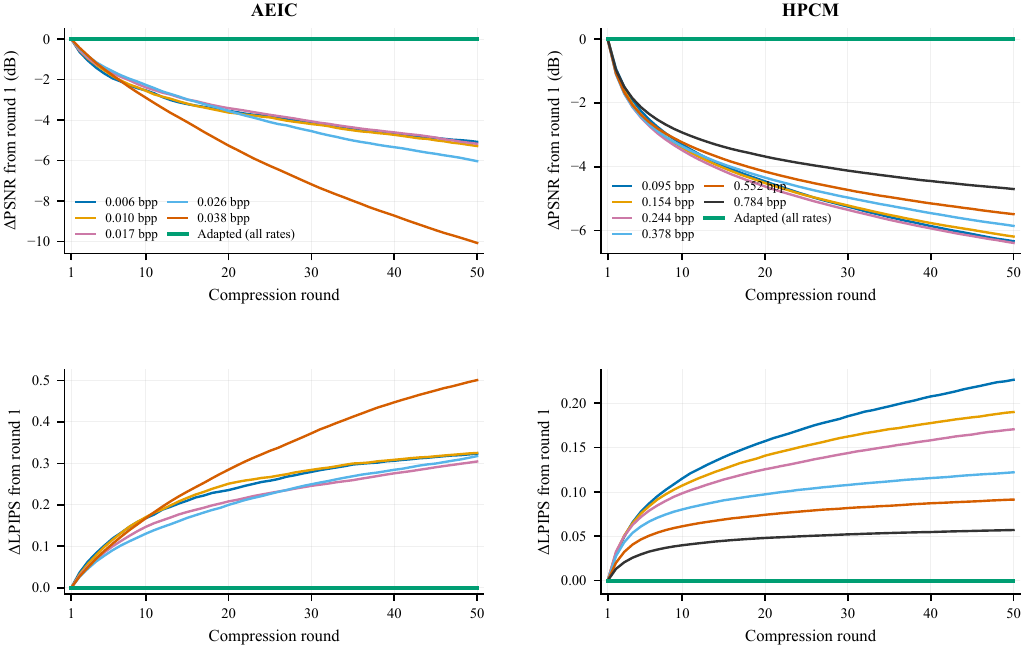}
\par\vspace{-0.1cm}\caption{\small\textbf{Recompression at all rates.} Each trajectory is measured, not extrapolated from one equality test. Zero drift is paired with the absolute quality and rate results in the main paper and detailed tables.}
\label{fig:all-drift}
\end{figure*}

Figure~\ref{fig:allmetrics} shows absolute quality across operating points, while Fig.~\ref{fig:all-drift} shows how quality changes over repeated compression. Table~\ref{tab:grand-all} gives the corresponding per-rate results and measured horizons.

\subsection{Clipped-Float and 8-bit Results}
\label{sec:png-full-rates}
The clipped-float tests preserve the initial files in all 120 AEIC chains (five rates) and all 96 HPCM chains (four rates, $\lambda\le.013$). Image equality follows from identical files and deterministic decoding; these tests do not include independent raw-tensor comparisons at every round.

Under 8-bit storage, AEIC preserves both the initial image and file in all 120 Kodak chains. HPCM preserves the image in all 144 chains and the initial file in 137. The other seven change their file once, at round 2, while retaining the first reconstruction. These are the seven chains whose round-1 decode the HPCM preset fails. Rerun with the escalation stages used for ELIC, Hyperprior, Mean-scale, and TCM instead of stabilization, with the stages unchanged, all seven pass at round 1 (two at stage 1, one at stage 2, four at the box step) and keep their initial file and image through 50 rounds, at 26--471\,s per decode. Their first-round PSNR differs from that of the stabilized chains by $-0.66$ to $+0.81$\,dB (mean $-0.04$\,dB). With these chains, the median decode over all 144 chains and 50 rounds stays at 9.9\,s, and the maximum rises from 253 to 471\,s. HPCM Kodak is a development set and the box step was designed after diagnosing these seven decodes, so the rerun is not a held-out test; the held-out evidence is the 870 frozen-setting chains above, all of which keep their first file and image.

ELIC's full procedure preserves the initial image and file in all 144 chains. Its preset repair succeeds in 141; escalation recovers the remaining three. The unchanged preset chains are combined with the measured escalation chains, with the development-set scope retained.

\begin{table*}[t]
\caption{\small\textbf{Per-rate 8-bit results.} PSNR is in dB. In (a), the initial compressed file is shared by the base and adapted decoders. All adapted images are constant through round 50, although seven files change once; PSNR$_{1,50}$ therefore has one value. In (b), $G=50$ for 41 CLIC images and 100 DIV2K images. The base PSNR change is relative to the base decoder's own first reconstruction; confidence intervals are 95\% percentile bootstrap intervals over images (10,000 resamples of the images, seed 20261114). Exact counts require agreement with round 1 at every round. CLIC totals 205/205 and DIV2K 500/500. All rates are payload rates; HPCM rates are inferred from the fixed framing, whereas the high-resolution rates are measured from parsed files.}\par\vspace{-0.1cm}
\label{tab:png-full-rates}
\centering\small
\setlength{\tabcolsep}{4pt}\renewcommand{\arraystretch}{1.08}
\textbf{(a) HPCM on Kodak: 24 images, 50 rounds, all six rates}\par\smallskip
\begin{tabular}{@{}rrrrrrrr@{}}
\toprule
$\lambda$ & \shortstack{Rate\\BPP$_1$} & \shortstack{Base\\PSNR$_1$} & \shortstack{Base\\PSNR$_{50}$} & \shortstack{Ours\\PSNR$_{1,50}$} & \shortstack{Ours\\BPP$_{50}$} & \shortstack{Image\\exact} & \shortstack{First-file\\exact}\\
\midrule
0.0018 & 0.09516 & 28.9133 & 22.5566 & 28.8756 & 0.09516 & 24/24 & 24/24\\
0.0035 & 0.15376 & 30.4489 & 24.1892 & 30.4277 & 0.15376 & 24/24 & 24/24\\
0.0067 & 0.24380 & 32.1220 & 25.6991 & 32.0952 & 0.24380 & 24/24 & 23/24\\
0.013 & 0.37784 & 33.9924 & 28.1213 & 33.9541 & 0.37784 & 24/24 & 24/24\\
0.025 & 0.55171 & 35.8512 & 30.2500 & 35.7036 & 0.55171 & 24/24 & 23/24\\
0.0483 & 0.78437 & 37.7480 & 32.5813 & 37.4583 & 0.78483 & 24/24 & 19/24\\
\bottomrule
\end{tabular}
\par\medskip
\textbf{(b) AEIC on CLIC and DIV2K: all five rates}\par\smallskip
\begin{tabular}{@{}llrrrrlr@{}}
\toprule
Dataset & Rate & BPP$_1$ & \shortstack{Ours\\BPP$_G$} & \shortstack{Base\\BPP$_G$} & \shortstack{Ours\\PSNR$_1$} & \shortstack{Base $\Delta$PSNR$_G$\\{[95\% CI]}} & \shortstack{Image/file\\exact}\\
\midrule
CLIC & ft2 & 0.03291 & 0.03291 & 0.04617 & 25.3606 & $-9.542\;[-10.599,-8.515]$ & 41/41\\
CLIC & ft4 & 0.02210 & 0.02210 & 0.02546 & 24.7229 & $-6.811\;[-7.812,-5.823]$ & 41/41\\
CLIC & ft8 & 0.01397 & 0.01397 & 0.01486 & 23.8458 & $-5.630\;[-6.451,-4.805]$ & 41/41\\
CLIC & ft16 & 0.00830 & 0.00830 & 0.00918 & 22.7222 & $-5.507\;[-6.158,-4.840]$ & 41/41\\
CLIC & ft32 & 0.00475 & 0.00475 & 0.00581 & 21.5677 & $-5.257\;[-5.938,-4.581]$ & 41/41\\
DIV2K & ft2 & 0.03676 & 0.03676 & 0.05221 & 23.7298 & $-9.268\;[-9.933,-8.624]$ & 100/100\\
DIV2K & ft4 & 0.02558 & 0.02558 & 0.02906 & 23.1023 & $-6.339\;[-6.997,-5.683]$ & 100/100\\
DIV2K & ft8 & 0.01661 & 0.01661 & 0.01739 & 22.2212 & $-5.231\;[-5.752,-4.710]$ & 100/100\\
DIV2K & ft16 & 0.00999 & 0.00999 & 0.01051 & 21.1249 & $-4.822\;[-5.228,-4.425]$ & 100/100\\
DIV2K & ft32 & 0.00569 & 0.00569 & 0.00647 & 19.9253 & $-4.659\;[-4.970,-4.348]$ & 100/100\\
\bottomrule
\end{tabular}
\end{table*}

Table~\ref{tab:png-full-rates} reports first-round quality, later quality and rate, and separate image and file equality. Its HPCM values were computed with each 8-bit image passed in memory; repeating the experiment as a file chain, which writes and reloads every compressed file and 8-bit image, gives identical files, bitstreams, symbols, and images in all 7,200 decodes. Small rate changes can remain when the image is exact but the initial file is not.

\begin{figure*}[!t]
\centering\includegraphics[width=\textwidth]{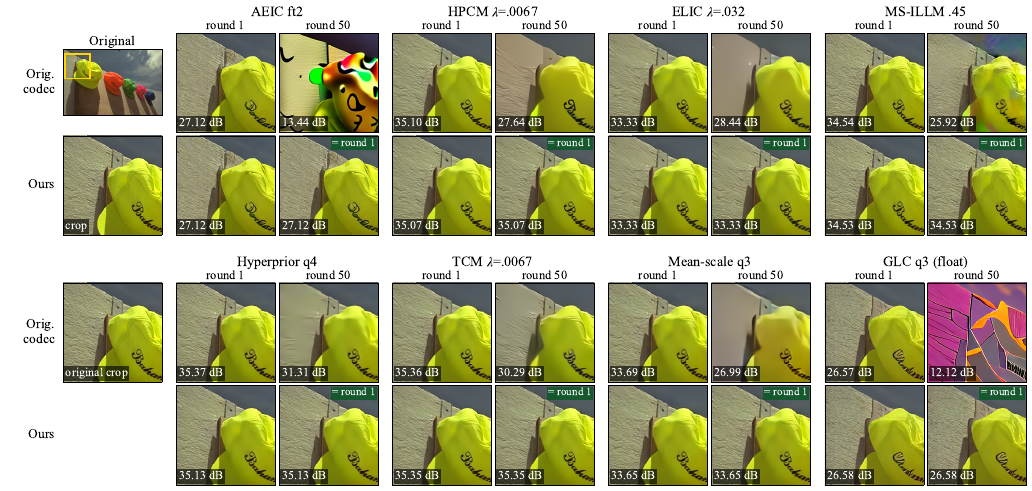}
\par\vspace{-0.1cm}\caption{\small\textbf{Recompression across eight pretrained codecs.} Kodak kodim03, rounds 1 and 50. For each codec, the top row shows the original codec and the bottom row its adapted version. All chains use 8-bit storage except GLC (float). PSNR (dB) is measured on the full image; the shared $192\times192$ crop was selected to highlight degradation in the original codecs. Each adapted round-50 image equals its round-1 image.}
\label{fig:codec-recompression-panel}
\end{figure*}
\FloatBarrier

\subsection{High-Resolution Results}
\label{sec:high-resolution}
Under the retained-canvas protocol, AEIC preserves its initial file in all 205 CLIC image--rate chains through 50 rounds. The DIV2K float experiment preserves the initial file in all 500 chains through 50 rounds, whereas the base AEIC decoder changes its file at round 2 in all 500 chains and loses $4.56$--$9.24$\,dB of mean PSNR by round 50.

For DIV2K 8-bit storage, the full method reaches 500/500 exact chains over 50 rounds. DIV2K was a development set for the escalation stages: the stage-0 preset passes in 493/500 chains, and escalation, which runs only after a failed acceptance check, recovers the remaining seven.

Table~\ref{tab:png-full-rates} gives the per-rate full-method results and Table~\ref{tab:adaptation-summary} the totals. The results concern the retained canvas and color metadata defined in Appendix~\ref{sec:datasets}.

\subsection{Comparisons with Existing Methods}
\label{sec:baseline-audit}
We separate measured 8-bit comparisons from native float results and published reference points.

\begin{figure}[t]
\centering\includegraphics[width=0.95\linewidth]{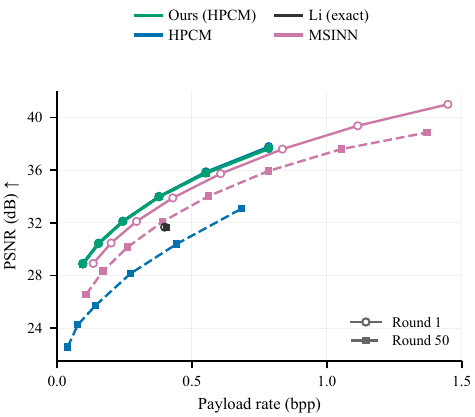}
\par\vspace{-0.1cm}\caption{\small \textbf{Quality retained through float-chain recompression.} Mean Kodak PSNR versus each round's actual payload rate. Colors identify methods; solid/open-circle and dashed/filled-square curves show rounds 1 and 50. The adapted HPCM curves overlap. Li's exact model supplies one point per round. Native float chains are scored after clamping; pixel-quantization checks are separate.}
\label{fig:idem-rd-float}
\end{figure}

\begin{table*}[t]
\caption{\small\textbf{Consistency methods on Kodak under 8-bit storage, initial rate $\approx.40$\,bpp.} All values measured by us; BPP is payload. PSNR in dB at rounds 1, 32 and 50; round-50 rates are .271 (HPCM), .378 (ours), .390 (MSINN) and 1.64 (Li) bpp. Training-free adaptation means applying consistency without retraining pretrained weights; ``--'' denotes an ordinary codec. $^{\ast}$Li's released q3 model is evaluated outside its native float chain, which keeps its image in all 24 chains (full-precision MSE at every round; pixel hashes on five images). \textbf{Bold}/\underline{underline}: best/second PSNR.}\par\vspace{-0.1cm}
\label{tab:png-methods}
\centering\footnotesize
\renewcommand{\arraystretch}{1.08}
\setlength{\tabcolsep}{3pt}
\begin{tabular*}{0.82\textwidth}{@{\extracolsep{\fill}}lcrrrr@{}}
\toprule
Method & \shortstack{Training-free\\adaptation} & BPP$_1$ & PSNR$_1$ & PSNR$_{32}$ & PSNR$_{50}$ \\
\midrule
HPCM & -- & 0.378 & \textbf{33.99} & 28.89 & 28.12 \\
HPCM + ours & Yes & 0.378 & \underline{33.95} & \textbf{33.95} & \textbf{33.95} \\
MSINN~\cite{msinn2025} (anchor 3) & No & 0.429 & 33.88 & \underline{31.93} & \underline{31.90} \\
Li (exact, q3)$^{\ast}$~\cite{rightinverse} & No & 0.399 & 31.66 & 12.94 & 10.75 \\
\bottomrule
\end{tabular*}

\end{table*}

Near $0.40$\,bpp initially, adapted HPCM retains $33.95$\,dB at round 50, compared with $28.12$\,dB for base HPCM and $31.90$\,dB for MSINN~\cite{msinn2025}. Between rounds 1 and 50, base HPCM loses $5.87$\,dB, MSINN $1.97$\,dB, and Li's exact model $20.91$\,dB, whereas the adapted decoder loses none after a $0.04$\,dB first-round cost. MSINN starts $0.051$\,bpp higher, and later rates are reported separately.

\begin{figure}[t]
\centering
\includegraphics[width=0.87\linewidth]{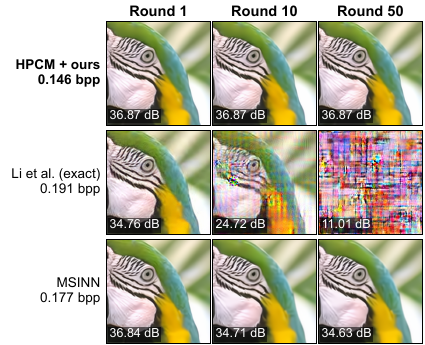}
\par\vspace{-0.1cm}\caption{\small\textbf{8-bit image recompression of \texttt{kodim23}, rounds 1, 10 and 50.} Full-image PSNR; rows give each method's initial payload bpp on this image (Kodak-mean rates $\approx.40$\,bpp). Ours (HPCM) stays exact, with identical images whether each round's 8-bit image is kept in memory or written to a file; Li et al.~\cite{rightinverse}, outside its native float chain, and MSINN~\cite{msinn2025} use 8-bit image files.}
\label{fig:png-idempotent}
\end{figure}

Li's released exact model~\cite{rightinverse} is also evaluated in its native float setting. Its image stability there and its degradation after 8-bit storage concern different protocols. Native-float image evidence combines full-precision quality checks with pixel hashes on the checked subset; it is not a full image-hash audit of every chain.

\begin{table*}[t]
\caption{\small\textbf{Published-result and native float-chain comparison on Kodak.} M: our native float-chain measurements; I: interpolation between initial-rate settings with fixed weights across rounds; P: published results, including values read from curves. HPCM rates are payload (12-byte framing excluded); MSINN reports entropy-stream bytes and writes no complete file. Published rates retain the source convention, which is not specified for the digitized Li/Helminger points. Li-near's rate and Li-exact's quality are read from Li et al.'s Fig.~3; $^{\dagger}$later exact-model values follow the published idempotence claim, not a measured run at this rate. Helminger's $32.04$\,dB is quoted in Li et al.'s Fig.~3(b), with its rate read from Fig.~3(a); dashes mark unavailable measured round-32/50 values at this point. The storage protocol of the published Li-near and Helminger curves is not specified sufficiently to identify them as our 8-bit protocol. Training-free adaptation has the same meaning as in Table~\ref{tab:png-methods}.}\par\vspace{-0.1cm}
\label{tab:published-comparison}
\centering\footnotesize
\renewcommand{\arraystretch}{1.08}
\setlength{\tabcolsep}{4pt}
\begin{tabular*}{\textwidth}{@{\extracolsep{\fill}}llcrrrr@{}}
\toprule
Method & Source & \shortstack{Training-free\\adaptation} & BPP$_1$ & PSNR$_1$ & PSNR$_{32}$ & PSNR$_{50}$ \\
\midrule
\multicolumn{7}{@{}l}{\textit{Initial rate $\approx .708$ bpp}} \\[1pt]
HPCM & M+I & -- & .708 & 37.16 & 32.75 & 32.20 \\
HPCM + ours & M+I & Yes & .708 & 37.03 & 37.03 & 37.03 \\
MSINN~\cite{msinn2025} & M+I & No & .708 & 36.56 & 34.89 & 34.89 \\
Li (exact)~\cite{rightinverse} & P & No & .708 & $\approx34.50$ & $\approx34.50^{\dagger}$ & $\approx34.50^{\dagger}$ \\
Li (near)~\cite{rightinverse} & P & No & .708 & 35.79 & 34.92 & 34.92 \\
\midrule
\multicolumn{7}{@{}l}{\textit{Nearby initial rates $.784$ / $.796$ bpp}} \\[1pt]
HPCM & M & -- & .784 & 37.78 & 33.59 & 33.08 \\
HPCM + ours & M & Yes & .784 & 37.64 & 37.64 & 37.64 \\
Helminger~\cite{flowcodec,rightinverse} & P & No & .796 & 32.04 & -- & -- \\
\bottomrule
\end{tabular*}

\end{table*}

Interpolation uses weights determined by the initial rates and holds them fixed across rounds:
\begin{equation}
P_g(r)=(1-w)P_g(r_\ell)+wP_g(r_h),\quad w=\frac{r-r_\ell}{r_h-r_\ell}.
\end{equation}
These values summarize neighboring checkpoints and do not represent newly encoded operating points. Published values retain their original rate and storage conventions. Later values derived from an idempotence claim are distinguished from executed recompression measurements.

The I2C comparison~\cite{cai2024i2c} in Fig.~\ref{fig:teaser}(b) uses its reported 32-round horizon. CDT's generation plot~\cite{stablecdt} does not provide the operating rate and uses dataset-averaged MSE for PSNR, so it does not support an equal-rate mean-PSNR ranking.

The separate $256\times256$ comparison uses Xu~\cite{xu2024}, plain ELIC, and adapted ELIC on matched inputs. Adapted ELIC preserves the initial image and file in all 24 Kodak and 41 CLIC chains through 50 rounds. Over the same rounds, Xu's method loses $8.15$\,dB (Kodak) and $7.73$\,dB (CLIC) and plain ELIC $3.33$\,dB on both sets, while adapted ELIC stays at $36.79$ and $36.82$\,dB, $0.05$ and $0.08$\,dB below plain ELIC's first round. These results are reported separately from native-resolution Kodak and high-resolution CLIC.

\FloatBarrier
\section{Ablations and Computational Cost}
\label{sec:ablations}
\subsection{Float Component Ablations}
\label{sec:float-ablations}
We evaluate three rates per codec and 24 Kodak images per rate through two executed rounds. The variants remove the target search (main-cell projection only), write the reconstructed latent $\hat y$ directly (wrapper only), replace PCA with a seeded random orthonormal basis, or use the wrapped encoder from the first generation.

\begin{table*}[t]
\centering\footnotesize\setlength{\tabcolsep}{3pt}
\caption{\small All ablations, Kodak, three rates per codec, 24 images per rate. Exactness compares round 2 with round 1. Failed searches remain in the denominator. Fail: images with a failed target search; --: the variant runs no target search. All variants keep neural weights frozen. Best results are in \textbf{bold} (ties are both bolded) and second-best are \underline{underlined}, per codec and rate in Equal, PSNR$_1$, PSNR$_2$, and LPIPS$_1$; Bpp, $\Delta R$, and Fail are not marked.}\par\vspace{-0.1cm}
\label{tab:ablation-full}
\begin{tabular}{lllrrrrrrr}
\toprule
Codec & Rate & Variant & Equal & PSNR$_1$ & PSNR$_2$ & LPIPS$_1$ & Bpp & $\Delta R$ (\%) & Fail\\\midrule
AEIC & ft2 & Full & \textbf{24/24} & \underline{23.297} & \textbf{23.297} & \textbf{0.1428} & 0.03792 & 0.00 & 0\\
AEIC & ft2 & No target search & 0/24 & \textbf{23.298} & 22.906 & \textbf{0.1428} & 0.03792 & 0.00 & --\\
AEIC & ft2 & Wrapper only & 0/24 & 23.159 & 23.007 & \textbf{0.1428} & 0.03792 & 0.00 & --\\
AEIC & ft2 & Random basis & \textbf{24/24} & 21.921 & 21.921 & 0.2803 & 0.03792 & 0.00 & 0\\
AEIC & ft2 & $E^{\prime}$ from round 1 & \textbf{24/24} & 23.292 & \underline{23.292} & 0.1432 & 0.03783 & -0.25 & 0\\
\addlinespace[3pt]
AEIC & ft8 & Full & \textbf{24/24} & \underline{22.045} & \underline{22.045} & \textbf{0.1979} & 0.01718 & 0.00 & 0\\
AEIC & ft8 & No target search & 0/24 & \underline{22.045} & 21.531 & \textbf{0.1979} & 0.01718 & 0.00 & --\\
AEIC & ft8 & Wrapper only & 0/24 & 21.988 & 21.521 & \textbf{0.1979} & 0.01718 & 0.00 & --\\
AEIC & ft8 & Random basis & \textbf{24/24} & 21.239 & 21.239 & 0.2912 & 0.01718 & 0.00 & 0\\
AEIC & ft8 & $E^{\prime}$ from round 1 & \textbf{24/24} & \textbf{22.056} & \textbf{22.056} & 0.1986 & 0.01724 & 0.37 & 0\\
\addlinespace[3pt]
AEIC & ft32 & Full & \textbf{24/24} & \underline{20.004} & \underline{20.004} & \underline{0.3114} & 0.00592 & 0.00 & 0\\
AEIC & ft32 & No target search & 0/24 & \underline{20.004} & 19.416 & \underline{0.3114} & 0.00592 & 0.00 & --\\
AEIC & ft32 & Wrapper only & 0/24 & 19.988 & 19.558 & \underline{0.3114} & 0.00592 & 0.00 & --\\
AEIC & ft32 & Random basis & \textbf{24/24} & 19.585 & 19.585 & 0.3843 & 0.00592 & 0.00 & 0\\
AEIC & ft32 & $E^{\prime}$ from round 1 & \textbf{24/24} & \textbf{20.023} & \textbf{20.023} & \textbf{0.3110} & 0.00596 & 0.69 & 0\\
\midrule
HPCM & $\lambda$ 0.013 & Full & \textbf{24/24} & \underline{33.975} & \textbf{33.975} & 0.1391 & 0.37784 & 0.00 & 0\\
HPCM & $\lambda$ 0.013 & No target search & 0/24 & \textbf{33.999} & 33.372 & 0.1392 & 0.37784 & 0.00 & --\\
HPCM & $\lambda$ 0.013 & Wrapper only & 0/24 & 33.658 & 33.415 & \underline{0.1388} & 0.37784 & 0.00 & --\\
HPCM & $\lambda$ 0.013 & Random basis & \textbf{24/24} & 17.448 & 17.448 & 0.6279 & 0.37784 & 0.00 & 0\\
HPCM & $\lambda$ 0.013 & $E^{\prime}$ from round 1 & \textbf{24/24} & 33.838 & \underline{33.838} & \textbf{0.1387} & 0.38912 & 2.98 & 0\\
\addlinespace[3pt]
HPCM & $\lambda$ 0.0035 & Full & \textbf{24/24} & \underline{30.432} & \textbf{30.432} & 0.2614 & 0.15375 & 0.00 & 0\\
HPCM & $\lambda$ 0.0035 & No target search & 0/24 & \textbf{30.453} & 29.761 & 0.2616 & 0.15376 & 0.00 & --\\
HPCM & $\lambda$ 0.0035 & Wrapper only & 0/24 & 30.361 & 30.077 & \underline{0.2613} & 0.15376 & 0.00 & --\\
HPCM & $\lambda$ 0.0035 & Random basis & \textbf{24/24} & 17.648 & 17.648 & 0.6279 & 0.15376 & 0.00 & 0\\
HPCM & $\lambda$ 0.0035 & $E^{\prime}$ from round 1 & \textbf{24/24} & 30.265 & \underline{30.265} & \textbf{0.2611} & 0.16773 & 9.09 & 0\\
\addlinespace[3pt]
HPCM & $\lambda$ 0.0018 & Full & \textbf{24/24} & 28.869 & \textbf{28.869} & \textbf{0.3245} & 0.09517 & 0.00 & 0\\
HPCM & $\lambda$ 0.0018 & No target search & 0/24 & \textbf{28.916} & 28.242 & 0.3254 & 0.09516 & 0.00 & --\\
HPCM & $\lambda$ 0.0018 & Wrapper only & 0/24 & \underline{28.870} & 28.567 & 0.3251 & 0.09516 & 0.00 & --\\
HPCM & $\lambda$ 0.0018 & Random basis & 22/24 & 16.728 & 16.727 & 0.7669 & 0.09516 & 0.00 & 2\\
HPCM & $\lambda$ 0.0018 & $E^{\prime}$ from round 1 & \textbf{24/24} & 28.685 & \underline{28.685} & \underline{0.3246} & 0.11067 & 16.30 & 0\\
\bottomrule\end{tabular}
\end{table*}

The full method preserves all 72 transitions per codec. Without the target search, and with the wrapper alone, which writes $\hat y$ as the decoder does when no target is accepted, no transition is preserved, and their round-2 PSNR is $0.39$--$0.67$ and $0.29$--$0.56$\,dB, respectively, below the full method's. The random basis preserves 142 of 144 transitions but costs up to $16.5$\,dB (HPCM, $\lambda=.013$: $33.975$ to $17.448$\,dB). Search is therefore necessary for consistency and the PCA basis for fidelity; neither suffices alone.

\subsection{8-bit Search and Repair Ablations}
\label{app:ablation-8bit}
We compare repair variants on seven ELIC decodes where the preset failed: Kodak at $\lambda=.45$, covering kodim10 round 1, kodim21 rounds 1--2, and kodim24 rounds 1--4. Each variant decodes the same stored files. The cases were selected after observing failures, so the counts measure hard-case recovery.

\begin{table}[t]
\centering\footnotesize\setlength{\tabcolsep}{3pt}
\caption{\small 8-bit repair variants on seven selected hard ELIC decodes. Pass: re-encoding the stored 8-bit image reproduces every main and hyper symbol. $\Delta$PSNR: difference from plain ELIC decoding of the same file, with ranges over all seven decodes. Times are maximum wall-clock decoding times on shared GPUs; no time limit was imposed.}
\label{tab:ablation-8bit}
\begin{tabular}{lrrr}
\toprule
Variant & Pass & $\Delta$PSNR (dB) & Max decode (s)\\\midrule
Preset (stage 0, update cap 30) & 0/7 & $-1.30$ to $-0.04$ & 141\\
No target update & 3/7 & $-0.21$ to $-0.02$ & 112\\
Update cap 120 & 5/7 & $-1.75$ to $-0.36$ & 25\\
Hyper-cell guard 0.10 & 3/7 & $-1.29$ to $-0.05$ & 17\\
Box step after stage 0 & 6/7 & $-0.84$ to $-0.02$ & 198\\
\bottomrule
\end{tabular}
\end{table}

Increasing the update cap recovers five cases, widening the hyper guard recovers three, and applying box search after the preset recovers six. The escalation ladder applies these changes in increasing order of cost, with a cap of 60 rather than 120, then the 0.10 guard, then the box step, so each runs only after the cheaper ones fail. Removing target updates recovers three with a smaller PSNR cost on this selected set. Across the 28 alternative decodes, passing the acceptance check coincides with an exact file re-encode and an identical re-decoded image.

On eight ordinary three-round chains, removing target updates retains exactness but takes $1.5$--$2.4\times$ longer. Target correction therefore also contributes to efficiency. Shared-GPU maximum times in the hard-case table are descriptive rather than a controlled speed comparison.

\subsection{First-Round Encoder Choice}
\label{sec:first-encoder-ablation}
The default procedure accepts the base codec's first file and uses $E'$ for later encodes. A separate AEIC ablation instead uses $E'$ from generation one.

\begin{table}[t]
\caption{\small\textbf{First-encoder ablation under 8-bit storage.} Each rate uses 24 Kodak images. Rates are measured payload BPP; ratios precede rounding. All chains retain the first file for 50 rounds; PSNR differences are rate-wise means (dB).}\par\vspace{-0.1cm}
\label{tab:png-first-encoder}
\centering\small
\setlength{\tabcolsep}{3pt}
\begin{tabular}{@{}lrrrr@{}}
\toprule
Rate & \shortstack{Base\\BPP$_1$} & \shortstack{Wrapped\\BPP$_1$} & Ratio & $\Delta$PSNR$_1$\\
\midrule
ft2 & 0.03792 & 0.04772 & 1.258 & $-0.073$\\
ft4 & 0.02635 & 0.03598 & 1.365 & $-0.108$\\
ft8 & 0.01718 & 0.02708 & 1.576 & $-0.098$\\
ft16 & 0.01042 & 0.02081 & 1.997 & $-0.198$\\
ft32 & 0.00592 & 0.01620 & 2.738 & $-0.183$\\
\bottomrule
\end{tabular}
\end{table}

All 120 chains remain file-exact through 50 rounds, but the first payload rate increases by factors of $1.258$--$2.738$, with rate-wise mean PSNR changes from $-0.073$ to $-0.198$\,dB. The cost reflects the 8-bit gain $c=10$: with the float gain $c=1$, the same choice changes AEIC's first rate by only $-0.25$ to $+0.69\%$ in the float ablations above. Using the wrapped encoder throughout therefore applies the same operator, $D_8\circ\Sigma\circ E'$, in every round and removes the first-generation distinction; the initial rate and quality cost arises under 8-bit storage, whereas in the float chain the first rate changes by less than 1\%.

\subsection{Computational Cost}
\label{app:cost-details}
The additional computation is at the receiver and is already present in regular decoding. It includes target search, correction, and any storage repair. NLBP adds the prior's sampling cost.

\begin{table*}[t]
\caption{\small\textbf{Decoding cost by protocol.} Seconds per image and decode; ranges over rates of per-rate medians. Timer F starts from the compressed file or code and includes entropy decoding; timer S starts from the decoded symbols, after entropy decoding. PNG storage is timed separately. Base is the original decoder under the same timer and GPU type; for AEIC with timer S on Kodak it is $0.143$\,s, measured in the float chain at all rates and in the 8-bit chain at ft2. Rows without $^{a}$ are as-run times on shared nodes. Rows with different timers or hardware are not a controlled speed comparison. Max is the slowest single decode. $^{a}$Dedicated timing run: synchronized, entropy decoding included, one RTX PRO 6000 after warm-up. $^{b}$Measured at $\lambda=.013$ only. $^{c}$Two of the 144 chains ($\lambda=.05$, kodim20 and kodim24) need the box stage at round 1. Their file is unchanged, so every round repeats this escalated decode (1,726--2,289\,s); 93 of these repeated decodes were not timed and enter the medians by rank, since each exceeds the median. In the slowest decode, 98.7\% of the time is the pixel search, which stages 0 and 1 each run to its 9,056-trial limit; stage 1 repeats stage 0's search because the best correction round does not change.}
\label{tab:runtime-placeholder}
\centering\small
\setlength{\tabcolsep}{3pt}
\begin{tabular}{@{}lllllrrrr@{}}
\toprule
Codec & Data & Protocol & Timer & GPU & Base & Ours r.\,1 & Ours r.\,$\geq$2 & Max \\
\midrule
AEIC & Kodak & Float$^{a}$ & F & PRO 6000 & 0.117--0.121 & 2.63--3.94 & 2.63--3.95 & 6.36 \\
AEIC & Kodak & Clipped float & S & B200 & 0.143 & 4.56--7.19 & 4.49--7.20 & 10.40 \\
AEIC & Kodak & 8-bit & S & B200 & 0.143 & 5.32--8.00 & 5.25--7.97 & 11.39 \\
AEIC & CLIC 2K & 8-bit & F & B200 & 8.25--8.37 & 22.3--27.6 & 20.3--26.2 & 69.9 \\
AEIC & DIV2K 2K & 8-bit & F & B200 & 8.56--8.64 & 22.0--29.1 & 21.4--28.2 & 228.9 \\
AEIC & DIV2K 2K & Float & S & B200 & 8.43--9.55 & 14.00--15.74 & 13.99--14.96 & 28.55 \\
HPCM & Kodak & Float$^{a}$ & F & PRO 6000 & 0.073--0.076 & 1.15--6.48 & 1.14--6.45 & 15.12 \\
HPCM & Kodak & Clipped float & S & B200 & 0.013$^{b}$ & 2.97--15.71 & 2.94--15.82 & 46.18 \\
HPCM & Kodak & 8-bit & F & B200 & 0.101--0.105 & 4.98--18.68 & 4.86--18.72 & 253.16 \\
ELIC & Kodak & Float$^{a}$ & F & PRO 6000 & 0.092--0.095 & 0.67--1.16 & 0.68--1.16 & 1.33 \\
ELIC & Kodak & 8-bit & F & B200 & 0.106--0.111 & 2.78--4.81 & 2.78--4.88 & 234.97 \\
MS-ILLM & Kodak & 8-bit & F & PRO 6000 & 0.067--0.069 & 2.78--3.12 & 2.52--2.86 & 4.10 \\
Hyperprior & Kodak & 8-bit & F & PRO 6000 & 0.030--0.056 & 1.27--2.32 & 1.01--2.05 & 7.05 \\
Mean-scale & Kodak & 8-bit & F & PRO 6000 & 0.030--0.056 & 1.64--2.62 & 1.38--2.36 & 43.78 \\
TCM & Kodak & 8-bit & F & PRO 6000 & 0.080--0.082 & 3.76--4.95 & 3.41--4.59 & 2289.03$^{c}$ \\
GLC & Kodak & Float$^{a}$ & F & PRO 6000 & 0.061--0.062 & 0.104--0.109 & 0.104--0.109 & 0.125 \\
\bottomrule
\end{tabular}

\end{table*}

The dedicated timing run uses one RTX PRO 6000, batch size one, and warm-up. As-run measurements use their recorded hardware and timer scopes. File-chain decode timers include compressed-file reading and entropy decoding; PNG storage is timed separately. Timers starting from in-memory symbols are marked S. Medians describe typical decoding; the maximum reflects rare escalated decodes. Because decoding is deterministic and stateless, a chain whose first decode escalates repeats that decode in every later round, and the medians count every such repetition.

For HPCM under 8-bit storage, the per-rate median ranges are $4.98$--$18.68$ seconds at round 1 and $4.86$--$18.72$ seconds at later rounds. The maximum is $253.16$ seconds over 7,200 decodes.

The timed HPCM run evaluates the 8-bit conversion on the GPU during search and retains one outer candidate variant, a change selected after inspecting an earlier execution of the same experiment. It reproduces that execution's outputs in all 7,200 decodes while reducing total decoding time 3.7$\times$, from 93.7 to 25.4 GPU-hours, and every measured decode finishes within 300 seconds. These timings come from a research implementation that prioritizes verifiable exactness over speed: every decode starts from the received file, keeps no state across rounds, and evaluates every search trial in the exact storage format. The HPCM result shows that output-preserving engineering can reduce this cost substantially; applying such changes to all codecs, and resuming the correction loop across escalation stages instead of restarting it, is left to future work.

\FloatBarrier
\section{Generative Decoding}
\label{sec:nlbp-full}
\subsection{NLBP Algorithm and Implementation}
\label{sec:nlbp-implementation}
The receiver starts from a consistent reconstruction $x_{\mathrm{base}}$ and chooses a frozen prior, noise level, and correction schedule. The received code remains fixed while the prior proposes image content. Table~\ref{tab:nlbp-configs} lists the configuration of every reported NLBP result.

\begin{algorithm}[t]
\caption{\small NLBP decoding}
\label{alg:nlbp}
\begin{minipage}{\linewidth}
\small
\textbf{Input:} symbols $s$, target $y^\star$, frozen diffusion model $\theta$.\par\smallskip
Initialize $x_\tau=\alpha_\tau x_{\mathrm{base}}+\beta_\tau\epsilon$.\par\smallskip
\textbf{For} $t\to t'$ in the sampling schedule:
\vspace{-0.1cm}
\[
\begin{aligned}
u_t&=\hat x_{0,\theta}(x_t,t),\quad y_t=E'(u_t),\\
v_t^+&=(1-\lambda_t)y_t+\lambda_t y^\star,\\
b_t^+&=b(u_t)+\kappa\lambda_t
[r_{\mathrm{ref}}-b(D_{\mathrm{ref}}(y_t))],\\
x_{t'}&=\alpha_{t'}W(v_t^+,b_t^+)+\beta_{t'}\tilde\epsilon_t.
\end{aligned}
\]
\textbf{End for}\par\smallskip
Apply final latent write and storage repair to obtain $x_{\mathrm{out}}$.\par\smallskip
\textbf{Return:} $x_{\mathrm{out}}$ and the check $\Sigma(E'(x_{\mathrm{out}}))=s$.
\end{minipage}
\end{algorithm}

\begin{figure*}[t]
\centering
\resizebox{\textwidth}{!}{%
\begin{tikzpicture}[font=\sffamily\small,
 block/.style={fill=codecgray,rounded corners=2pt,align=center,inner sep=7pt},
 arr/.style={-{Stealth[length=1.8mm,width=1.2mm]},draw=black!65,line width=.7pt}]
\path[use as bounding box](-.2,-1.25)rectangle(18.7,1.1);
\node[block] (base) at (1.15,0) {Consistent decode\\$x_{\rm base}$};
\node[block] (noise) at (3.95,0) {Add noise\\at start $\tau$};
\node[block,fill=codecblue,text=white] (prior) at (7,0) {Chosen frozen prior\\clean proposal $u$};
\node[block,fill=codeclight] (bp) at (10.55,0) {NLBP toward $y^\star$\\Algorithm~\ref{alg:nlbp}};
\node[block] (sched) at (13.55,0) {Scheduler};
\node[block,fill=codeclight] (out) at (16.8,0) {Final exact write\\and full code check};
\draw[arr](base)--(noise);\draw[arr](noise)--(prior);\draw[arr](prior)--(bp);
\draw[arr](bp)--(sched);\draw[arr](sched)--(out);
\draw[arr](sched.south)--(13.55,-1.05)--(7,-1.05)--(prior.south);
\node[fill=white,font=\sffamily\scriptsize] at (10.4,-1.05) {repeat selected denoising / projection steps};
\node[font=\sffamily\scriptsize,text=codecblue] at (10.55,.85) {same received code throughout};
\end{tikzpicture}}
\par\vspace{-0.1cm}\caption{\small\textbf{Generative decoding at fixed code.} The prior proposes complementary content and NLBP enforces the target representative. The start noise, correction coordinates, strength, and schedule control the reconstruction. For a latent-space prior, decode its clean proposal before the image-space correction and account for the subsequent re-encoding. A full final write is followed by an operational code check. At zero start noise, the designated baseline bypasses the loop.}
\label{fig:method}
\end{figure*}
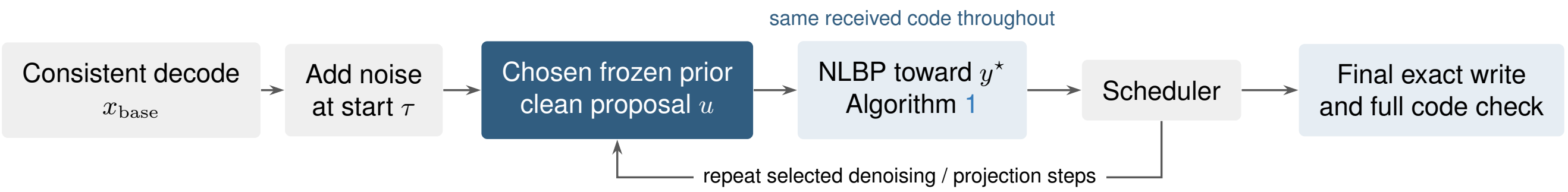

The clean estimate $u_t$, coefficients, and effective noise follow the prior's sampling convention. For latent-space priors, the clean proposal is decoded before image-space correction. At zero starting noise, the designated baseline bypasses sampling.

The reference is $r_{\mathrm{ref}}=b(D_{\mathrm{ref}}(v_{\mathrm{ref}}))$, with $v_{\mathrm{ref}}\in\{\hat y,y^\star\}$ as configured. Setting $\kappa=0$ preserves the proposal's complementary coordinates; nonzero $\kappa$ incorporates the reference decoder.

Before the final write, an optional bounded optimization adjusts $b$ to reduce
\begin{equation}
\|W(y^\star,b)-u_{\mathrm{final}}\|_2^2,
\end{equation}
retaining the best evaluated iterate. The 8-bit repair may select a separate target $y^\star_8$ for the same symbols. Intermediate corrections do not replace the final stored-image check. Failed outputs remain uncertified, and the procedure does not establish exact posterior sampling.

\subsection{FFHQ Evaluation}
\label{sec:ffhq-evaluation}
Both generative methods use the same ELIC checkpoints and the unconditional \texttt{ffhq\_10m} prior. Configurations are selected on 100 tuning images and evaluated on 1,000 disjoint test images at five rates.

\begin{table*}[t]
\caption{\small\textbf{Generative decoding on FFHQ at five operating points.} These are first-decoding results on the same 1,000 images, with common ELIC checkpoints and the shared pretrained prior for the two generative decoders. Each block uses its exported common measured payload rate; this does not independently certify byte-identical inputs. Scores use saved 8-bit images. Our NLBP rows use the configuration selected on the tuning images and frozen before testing; the ELIC and regular rows come from an earlier export that truncated to 8 bits (rounding raises ELIC's PSNR by at most 0.09\,dB). FID/KID use 16,000 nonoverlapping $64\!\times\!64$ patches, so their absolute values should not be compared with the Kodak table. \textbf{Bold} and \underline{underlined} scores denote the best and second distinct displayed values within each rate block, with ties sharing rank. The table separates distortion fidelity, perceptual quality, and decoding method. Xu's sampler is unseeded, so its scores differ slightly between runs (LPIPS $0.0295$ here and $0.0294$ in the 50-round run at the same rate).}\par\vspace{-0.1cm}
\label{tab:nlbp-quality}
\centering
\footnotesize
\setlength{\tabcolsep}{7pt}
\renewcommand{\arraystretch}{1.08}
\begin{tabular}{@{}lrrrrrr@{}}
\toprule
Decoding method & PSNR $\uparrow$ & MS-SSIM $\uparrow$ & LPIPS $\downarrow$ & DISTS $\downarrow$ & FID$_{64}$ $\downarrow$ & KID$_{64}\!\times\!10^3$ $\downarrow$ \\
\midrule
\multicolumn{7}{@{}l}{\textit{ELIC checkpoint 1: 0.07974 bpp}} \\[1pt]
ELIC & \textbf{28.42} & \textbf{0.9482} & 0.2480 & 0.2444 & 91.88 & 70.62 \\
ELIC + ours (regular) & \textbf{28.42} & \textbf{0.9482} & 0.2481 & 0.2444 & 91.58 & 70.29 \\
ELIC + ours (NLBP) & \underline{26.84} & \underline{0.9266} & \underline{0.1207} & \underline{0.1359} & \underline{9.90} & \underline{4.02} \\
Xu et al.~\cite{xu2024} & 26.53 & 0.9219 & \textbf{0.1174} & \textbf{0.1271} & \textbf{5.85} & \textbf{1.49} \\
\midrule
\multicolumn{7}{@{}l}{\textit{ELIC checkpoint 2: 0.11513 bpp}} \\[1pt]
ELIC & \textbf{29.88} & \textbf{0.9644} & 0.1943 & 0.2142 & 76.21 & 57.33 \\
ELIC + ours (regular) & \textbf{29.88} & \textbf{0.9644} & 0.1943 & 0.2142 & 75.71 & 56.76 \\
ELIC + ours (NLBP) & \underline{28.22} & \underline{0.9482} & \underline{0.0917} & \underline{0.1209} & \underline{8.59} & \underline{3.21} \\
Xu et al.~\cite{xu2024} & 27.92 & 0.9462 & \textbf{0.0846} & \textbf{0.1100} & \textbf{5.20} & \textbf{1.35} \\
\midrule
\multicolumn{7}{@{}l}{\textit{ELIC checkpoint 3: 0.16438 bpp}} \\[1pt]
ELIC & \textbf{31.36} & \textbf{0.9751} & 0.1498 & 0.1867 & 64.11 & 48.24 \\
ELIC + ours (regular) & \textbf{31.36} & \textbf{0.9751} & 0.1498 & 0.1866 & 63.50 & 47.56 \\
ELIC + ours (NLBP) & \underline{29.40} & \underline{0.9619} & \underline{0.0692} & \underline{0.1067} & \underline{7.40} & \underline{2.56} \\
Xu et al.~\cite{xu2024} & 29.19 & 0.9613 & \textbf{0.0604} & \textbf{0.0949} & \textbf{4.75} & \textbf{1.30} \\
\midrule
\multicolumn{7}{@{}l}{\textit{ELIC checkpoint 4: 0.37177 bpp}} \\[1pt]
ELIC & \textbf{34.82} & \textbf{0.9896} & 0.0747 & 0.1256 & 37.86 & 28.83 \\
ELIC + ours (regular) & \underline{34.81} & \textbf{0.9896} & 0.0747 & 0.1256 & 36.75 & 27.65 \\
ELIC + ours (NLBP) & 32.83 & \underline{0.9845} & \underline{0.0303} & \underline{0.0709} & \underline{5.16} & \underline{1.69} \\
Xu et al.~\cite{xu2024} & 32.72 & 0.9836 & \textbf{0.0295} & \textbf{0.0661} & \textbf{4.11} & \textbf{1.29} \\
\midrule
\multicolumn{7}{@{}l}{\textit{ELIC checkpoint 5: 0.66081 bpp}} \\[1pt]
ELIC & \textbf{37.44} & \textbf{0.9947} & 0.0387 & 0.0841 & 24.19 & 19.50 \\
ELIC + ours (regular) & \underline{37.39} & \textbf{0.9947} & 0.0387 & 0.0840 & 22.73 & 18.01 \\
ELIC + ours (NLBP) & 35.71 & \underline{0.9932} & \underline{0.0187} & \textbf{0.0507} & \underline{5.29} & \underline{2.47} \\
Xu et al.~\cite{xu2024} & 35.46 & 0.9914 & \textbf{0.0186} & \underline{0.0512} & \textbf{4.02} & \textbf{1.56} \\
\bottomrule
\end{tabular}

\end{table*}

Our NLBP decoder has higher PSNR than Xu~\cite{xu2024} at all five rates, by $0.11$--$0.31$\,dB, while Xu has lower FID, by $1.05$--$4.05$, and lower KID. At $0.1644$\,bpp, NLBP lowers FID from $63.50$ to $7.40$ ($8.6\times$) at a PSNR cost of $1.96$\,dB relative to regular adapted decoding.

The final 8-bit re-encoding check preserves the target symbols for 1,000/1,000 images at each of the first four rates and 990/1,000 at the highest. The ten failures remain in the quality averages. These are first-output checks, separate from the repeated-compression experiments below. The check runs in the decoder, so each failure is detected at decode time. The 50-round experiment below uses the fourth rate ($.3718$\,bpp), at which all 1,000 checks pass.

\subsection{Reconstruction Control and Quality Trade-offs}
\label{sec:decoder-controls}
The prior and its influence can be chosen after encoding. Starting noise controls how far sampling moves from the consistent baseline; correction strength and complementary-coordinate updates provide additional control.

\begin{figure}[t]
\centering\includegraphics[width=\linewidth]{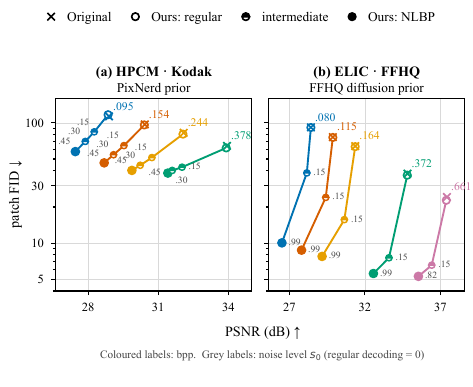}
\par\vspace{-0.1cm}\caption{\small\textbf{Decoder choices across rates.} At each bpp, a colored path runs from regular adapted decoding ($\circ$) through measured intermediate noise starts (half-filled) to an NLBP setting ($\bullet$); crosses mark the original codec. Grey numbers give each start's noise level $\omega=\sqrt{1-\bar\alpha_\tau}/(\sqrt{\bar\alpha_\tau}+\sqrt{1-\bar\alpha_\tau})$, the weight of $\epsilon$ in $x_\tau$ rescaled to $(1-\omega)\,x_{\rm base}+\omega\,\epsilon$. (a) HPCM with PixNerd on all 24 Kodak images. (b) ELIC with an FFHQ diffusion prior on 1,000 test images. All points at a rate share one bitstream. PSNR is linear, patch FID logarithmic; FID protocols differ.}
\label{fig:rate}
\end{figure}

Figure~\ref{fig:rate} varies starting noise while holding the other settings fixed along each path. HPCM uses PixNerd~\cite{pixnerd2026} on Kodak, and ELIC uses an FFHQ diffusion prior. Every point is measured, and each path shares one transmitted representation per image. The datasets use different FID protocols.

The ELIC paths use the earlier control configuration; Table~\ref{tab:nlbp-quality} reports the separately tuned final configuration. Their endpoint scores are therefore identified by configuration. In the teaser figure, the ELIC curve differs from the final comparison in its intermediate points, which use the untuned start $\omega=.15$; its endpoints use the final configuration.

\begin{figure*}[t]
\centering\includegraphics[width=\textwidth]{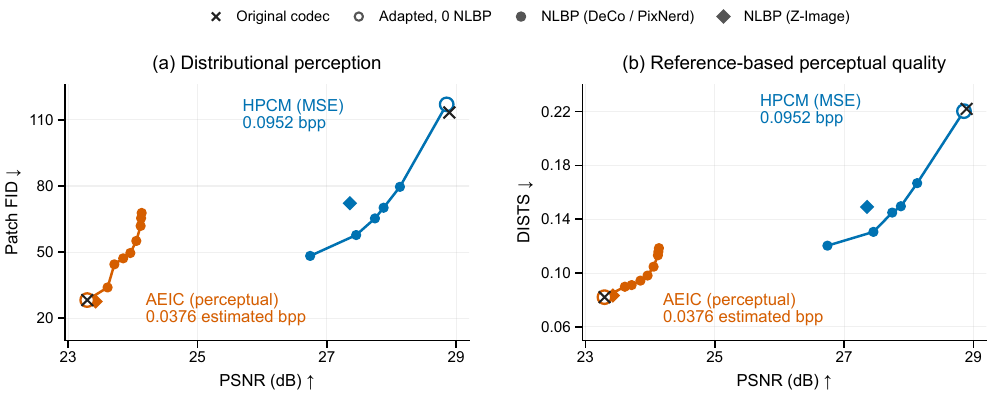}
\par\vspace{-0.1cm}\caption{\small\textbf{All measured fixed-code decoder choices.} Each codec has a fixed set of transmitted representations over all 24 Kodak images. Rates differ between codecs. This figure shows the complete AEIC and HPCM control sweeps. Lines connect selected settings rather than asserting a continuous or monotonic response. Separate priors are identified separately.}
\label{fig:decoder-control}
\end{figure*}

Figure~\ref{fig:decoder-control} includes changes in noise, correction strength, refinement, and prior. Its lines connect evaluated settings and do not imply a one-factor or monotonic response. Odd-numbered Kodak images were used for selection and even-numbered images for checking; the figure pools all 24.

The HPCM experiments use \texttt{MCG-NJU/PixNerd-XXL-P16-T2I}, with additional Z-Image settings~\cite{zimage} and, in the generator-choice comparison, FLUX.2 [dev]. AEIC uses DeCo-512~\cite{deco2026} and additional Z-Image settings. These examples demonstrate receiver-side choices at fixed per-codec representations; they do not establish an equal-rate ranking across codecs. Some settings introduce fine mesh artifacts.

\subsection{Recompression and Qualitative Examples}
\label{sec:nlbp-recompression-results}
Figure~\ref{fig:nlbp-recompression} follows all 1,000 FFHQ test images of Table~\ref{tab:ffhq-main} through 50 rounds of 8-bit recompression. NLBP preserves its round-1 file at every round, retaining $32.83$\,dB and FID$_{64}$ $5.16$. By round 50, ELIC falls from $34.87$ to $31.10$\,dB and Xu from $32.72$ to $24.88$\,dB, while their FID$_{64}$ rises from $37.85$ to $59.30$ and from $4.11$ to $36.38$. Xu's first-decoding FID advantage thus becomes a $7.0\times$ deficit by round 50.

\begin{figure*}[t]
\centering\includegraphics[width=\textwidth]{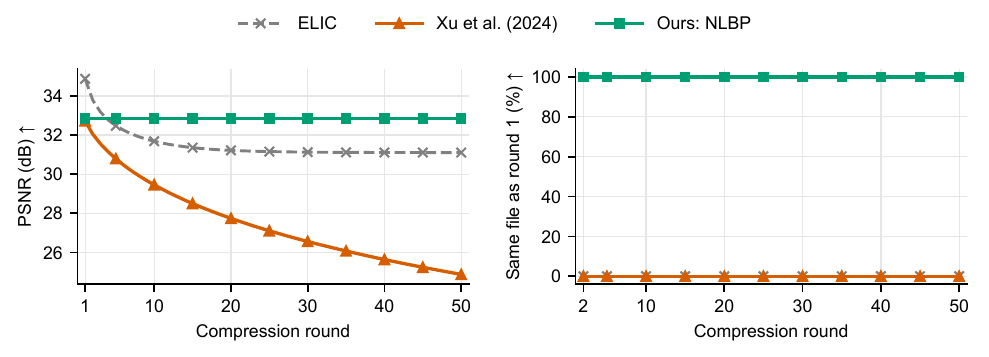}
\par\vspace{-0.1cm}\caption{\small\textbf{Generative decoding under 50 rounds of 8-bit recompression.} All 1,000 FFHQ test images, ELIC at $.3718$\,bpp (Table~\ref{tab:ffhq-main}). Left: mean PSNR of each round's 8-bit image. Right: share of images whose bitstream equals round 1. Ours (NLBP) keeps every round-1 file; ELIC and Xu (unseeded) keep none after round 1.}
\label{fig:nlbp-recompression}
\end{figure*}

Bitstream equality is checked at every round, and saved images at rounds 1, 2, 10, 25 and 50. NLBP and ELIC decode the same round-1 file for all 1,000 images; Xu's round-1 file is identical for 987 images and differs by at most 4 bytes for 13.

A separate experiment runs 100 rounds of 8-bit recompression on five images at the highest FFHQ rate. The images were selected for the largest DISTS improvement over Xu in the earlier comparison. It uses the earlier control configuration at this rate (start noise $0.82$), not the final FFHQ configuration ($0.67$).

\begin{figure*}[t]
\centering\includegraphics[width=0.85\textwidth]{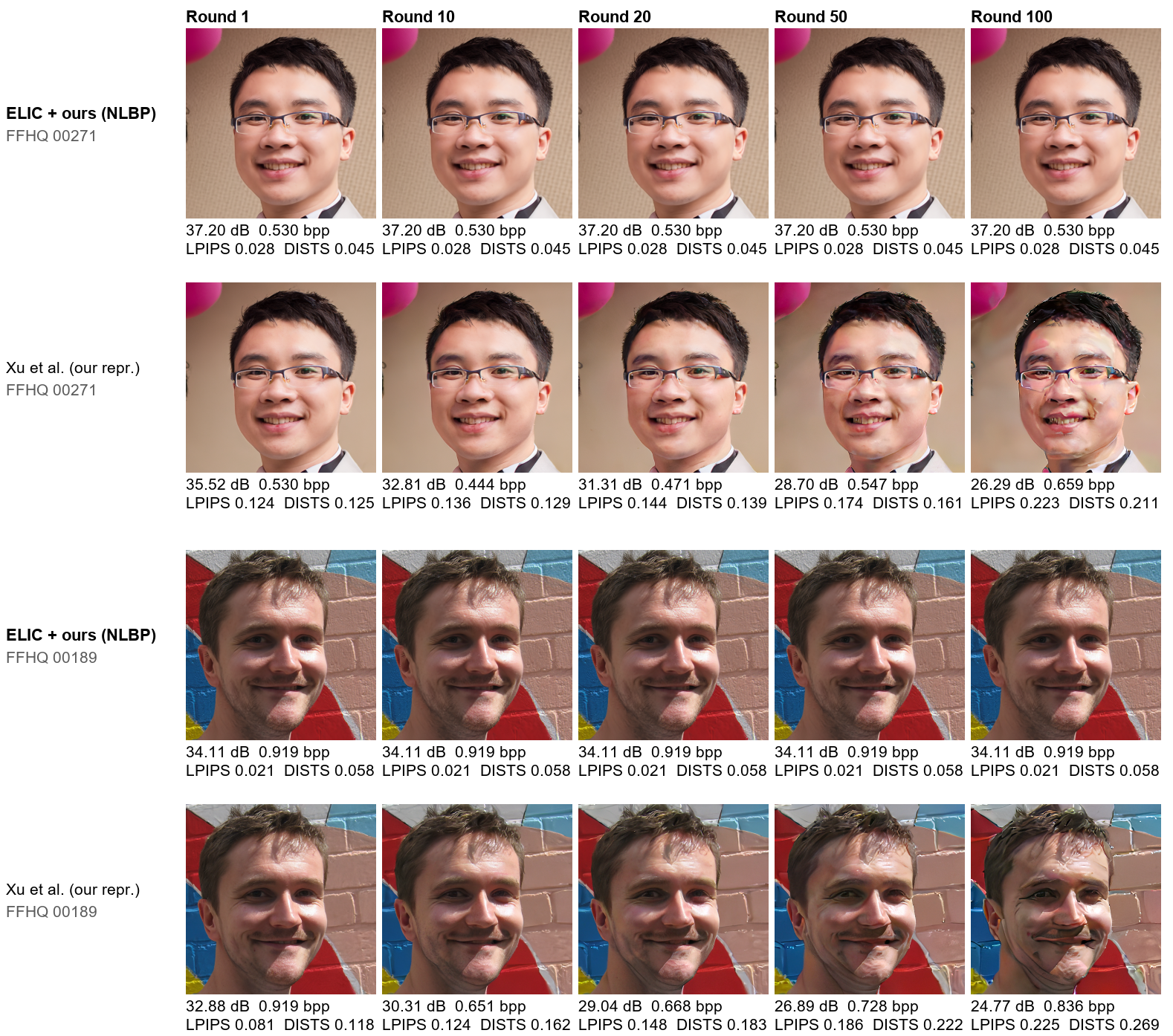}
\par\vspace{-0.1cm}\caption{\small\textbf{100 rounds of 8-bit image recompression on FFHQ: ELIC + ours (NLBP) vs.\ Xu et al.~\cite{xu2024}.} Two of the five images of the illustrative 100-round experiment at the highest FFHQ rate; the images were selected for the largest DISTS improvement over Xu in the earlier comparison, so the example is not representative. Each tile is the saved 8-bit image from that round; labels give PSNR, LPIPS and DISTS against the original and the payload rate of that round's file (for Xu, of a plain-ELIC re-encoding of the round's input 8-bit image, a proxy for the file Xu decoded). Ours reproduces its round-1 image and file at every round. Xu is our reproduction with its code unchanged; from round 2 its input is ELIC's encoding of its previous output, so its rate changes across rounds.}
\label{fig:nlbp-xu100}
\end{figure*}

Both regular adapted decoding and NLBP retain the initial file and image across all 495 subsequent image--round pairs. Xu's displayed file rates use a plain-ELIC re-encoding of each round's input image as a proxy for the decoded file. The example is illustrative and does not replace the 1,000-image evaluation or its reported failures.

\FloatBarrier
\clearpage\onecolumn
\section*{Detailed Quality, Rate, and Runtime}
\phantomsection\label{sec:grand-results}
Table~\ref{tab:grand-all} gives detailed quality, rate, and runtime results for the evaluated recompression protocols. Definitions and scoring conventions are in Appendix~\ref{sec:full-results}; timing scopes are in Appendix~\ref{app:cost-details}. Preset and escalated configurations are identified separately. Missing measurements are shown as dashes.

\begingroup\fontsize{6}{6.9}\selectfont\setlength{\tabcolsep}{1.1pt}\renewcommand{\arraystretch}{1.15}
\noindent Quality and rate cells show round 1 above the last measured round. Rows pair the base codec with codec + ours at the same setting.
B: base codec; F: frozen adaptation; T: trained consistency method. Image exact and File exact count chains equal to round 1 throughout the measured rounds.
File exact refers to the complete compressed file, not the PNG. A dash means unavailable; N/A means the evaluation has no complete compressed-file representation.
PSNR drift is last minus first (dB), using paired measurements or source means before table rounding. BPP counts payload and retained colour metadata, excluding container framing; GLC rate is estimated.
\par\smallskip
The Float block retains each codec's native decoder operations. Native-clipped base decoders also serve as the Clipped-float baselines.
$^{\dagger}$ Image equality is inferred from identical files and deterministic decoding, without independent all-round raw-float comparison.
ELIC/GLC float image counts use the in-run flag at all rounds and independent saved-image checks at rounds 2, 10, 50.
GLC q3 quality uses the same 23 completed images at both endpoints; exact counts retain all 24 (one base chain fails at round 44).
Li's native float stream changes once, at round 2, while its image stays constant in all 24 chains (full-precision MSE at every round; pixel hashes on five images). Native entropy-stream equality is not a complete-file test; missing MSINN image counts remain dashes.
\par\smallskip
$^{*}$ Full method after post-hoc escalation: 3 ELIC chains (50 rounds), 7 DIV2K chains (50 rounds). Other chains are unchanged by the code path and were not rerun.
Unavailable full-cohort perceptual scores and timing after substitution remain dashes.
$^{\mathrm{ND}}$ Xu uses an unseeded sampler and is not bitwise deterministic across nodes.
\par\smallskip
Timing sources: K, separate synchronized Kodak float run (round-1 medians); S, clipped-float/AEIC 8-bit (as-run medians);
E, ELIC/GLC float (all-round medians); R, adapted HPCM and adapted ELIC 8-bit file chains except ELIC $\lambda=.45$ (round-1 medians);
P, other 8-bit file chains, including base HPCM and adapted ELIC $\lambda=.45$ (all-round medians); M, scorer exports for MS-ILLM, Hyperprior, Mean-scale and TCM (all-round medians; for TCM $\lambda=.05$, 93 untimed box-stage decodes, each slower than the median, are counted by rank);
X, matched $256^2$ Xu protocol (round-1 medians where available); C, native Li/MSINN; L, Li clipped float;
D/H, DIV2K float and CLIC/DIV2K 8-bit canvas runs. Hardware and timing scopes differ; these are not a common speed benchmark.
Kodak FID/KID uses 2,304 patches of size $64^2$ (GLC q3: 2,208); X uses 384/656 patches; high-resolution H uses $256^2$ patches.
K-row AEIC first-round FID/KID retain the existing reference values.
\par\medskip
\begin{longtable}{@{}>{\raggedright\arraybackslash}p{63.5pt}>{\raggedright\arraybackslash}p{30pt}>{\centering\arraybackslash}p{8.5pt}>{\centering\arraybackslash}p{14.25pt}>{\centering\arraybackslash}p{21pt}>{\centering\arraybackslash}p{26.75pt}>{\centering\arraybackslash}p{25pt}>{\centering\arraybackslash}p{33.5pt}>{\centering\arraybackslash}p{23.5pt}>{\centering\arraybackslash}p{23.5pt}>{\centering\arraybackslash}p{24.25pt}>{\centering\arraybackslash}p{28.5pt}>{\centering\arraybackslash}p{26.75pt}>{\centering\arraybackslash}p{28.5pt}>{\centering\arraybackslash}p{28.5pt}>{\centering\arraybackslash}p{31.75pt}>{\centering\arraybackslash}p{15pt}@{}}
\caption{\small Detailed results by recompression protocol.}\label{tab:grand-all}\\[-0.1cm]
\toprule
Method & Setting & Tr. & $n$ & Rounds & BPP & PSNR$\uparrow$ & MS-SSIM$\uparrow$ & LPIPS$\downarrow$ & DISTS$\downarrow$ & FID$\downarrow$ & KID$\downarrow$ & \shortstack{Enc./Dec.\\(s)} & \shortstack{Image\\exact} & \shortstack{File\\exact} & \shortstack{PSNR drift\\(dB)} & Src. \\
\midrule\endfirsthead
\multicolumn{17}{l}{\textit{Detailed results by recompression protocol (continued)}}\\
\toprule
Method & Setting & Tr. & $n$ & Rounds & BPP & PSNR$\uparrow$ & MS-SSIM$\uparrow$ & LPIPS$\downarrow$ & DISTS$\downarrow$ & FID$\downarrow$ & KID$\downarrow$ & \shortstack{Enc./Dec.\\(s)} & \shortstack{Image\\exact} & \shortstack{File\\exact} & \shortstack{PSNR drift\\(dB)} & Src. \\
\midrule\endhead
\midrule\multicolumn{17}{r}{\textit{Continued on next page}}\\\endfoot
\bottomrule\endlastfoot
\addlinespace[3pt]\multicolumn{17}{l}{\textbf{Float --- Kodak}}\\*\addlinespace[2pt]
AEIC & ft2 & B & 24 & 1$\to$50 & \shortstack{0.0379\\0.0558} & \shortstack{23.30\\13.23} & \shortstack{0.8324\\0.4581} & \shortstack{0.143\\0.644} & \shortstack{0.082\\0.347} & \shortstack{28.20\\142.63} & \shortstack{0.00143\\0.07259} & \shortstack{0.028\\0.120} & 0/24 & 0/24 & -10.067 & K \\* %
AEIC + ours & ft2 & F & 24 & 1$\to$50 & \shortstack{0.0379\\0.0379} & \shortstack{23.30\\23.30} & \shortstack{0.8324\\0.8324} & \shortstack{0.143\\0.143} & \shortstack{0.082\\0.082} & \shortstack{28.27\\28.27} & \shortstack{0.00145\\0.00145} & \shortstack{0.029\\3.942} & 24/24 & 24/24 & 0.000 & K \\ %
AEIC & ft4 & B & 24 & 1$\to$50 & \shortstack{0.0264\\0.0304} & \shortstack{22.77\\16.75} & \shortstack{0.8025\\0.5970} & \shortstack{0.164\\0.481} & \shortstack{0.088\\0.241} & \shortstack{29.04\\80.55} & \shortstack{0.00132\\0.02377} & \shortstack{0.028\\0.118} & 0/24 & 0/24 & -6.027 & K \\* %
AEIC + ours & ft4 & F & 24 & 1$\to$50 & \shortstack{0.0264\\0.0264} & \shortstack{22.77\\22.77} & \shortstack{0.8025\\0.8025} & \shortstack{0.164\\0.164} & \shortstack{0.088\\0.088} & \shortstack{29.04\\29.04} & \shortstack{0.00134\\0.00134} & \shortstack{0.029\\3.550} & 24/24 & 24/24 & 0.000 & K \\ %
AEIC & ft8 & B & 24 & 1$\to$50 & \shortstack{0.0172\\0.0175} & \shortstack{22.05\\16.88} & \shortstack{0.7598\\0.5883} & \shortstack{0.198\\0.502} & \shortstack{0.101\\0.257} & \shortstack{31.46\\80.29} & \shortstack{0.00169\\0.02372} & \shortstack{0.028\\0.117} & 0/24 & 0/24 & -5.167 & K \\* %
AEIC + ours & ft8 & F & 24 & 1$\to$50 & \shortstack{0.0172\\0.0172} & \shortstack{22.04\\22.04} & \shortstack{0.7598\\0.7598} & \shortstack{0.198\\0.198} & \shortstack{0.101\\0.101} & \shortstack{31.46\\31.46} & \shortstack{0.00168\\0.00168} & \shortstack{0.029\\2.844} & 24/24 & 24/24 & 0.000 & K \\
\end{longtable}
\endgroup
\clearpage

\begin{table*}[t]
\caption{\small\textbf{Forced failures and trained consistency codecs over 50 rounds of recompression.} Kodak, 24 images; mean PSNR (dB) at rounds 1 and 50 and its change $\Delta$, computed before rounding. In the two forced arms, each decode returns a failing candidate whenever one exists: the closest candidate (fewest mismatched symbols; in clipped float, the last range-repair iterate before the accepted one) or the unrepaired output ($W(\hat y,b)$ in float; $W(y^\star,b)$ before range repair or 8-bit correction otherwise). Forced: decodes that returned a failing candidate, of 1,200 per arm. Rows of the base codecs, the full method, and the trained codecs repeat the full-results table; bpp is the first-round rate.}
\label{tab:forced-failures}
\centering\small
\setlength{\tabcolsep}{5pt}
\begin{tabular}{@{}lllrrrr@{}}
\toprule
Codec, rate & Decoder & bpp & PSNR$_1$ & PSNR$_{50}$ & $\Delta$ & Forced\\
\midrule
\multicolumn{7}{@{}l}{\textit{Float}}\\[1pt]
AEIC, ft8 & base codec & 0.017 & 22.05 & 16.88 & $-5.17$ & --\\
 & ours, full method & 0.017 & 22.04 & 22.04 & 0.00 & --\\
 & ours, closest candidate & 0.017 & 22.01 & 20.41 & $-1.59$ & 1,200\\
 & ours, unrepaired output & 0.017 & 21.99 & 20.74 & $-1.25$ & 255\\
\midrule
\multicolumn{7}{@{}l}{\textit{Clipped float}}\\[1pt]
HPCM, $\lambda{=}.013$ & base codec & 0.378 & 34.01 & 28.15 & $-5.86$ & --\\
 & ours, full method & 0.378 & 33.97 & 33.97 & 0.00 & --\\
 & ours, closest candidate & 0.378 & 33.97 & 33.97 & 0.00 & 1,200\\
 & ours, unrepaired output & 0.378 & 33.97 & 33.76 & $-0.22$ & 1,200\\
Li (exact), q3 & trained & 0.399 & 31.67 & 16.11 & $-15.56$ & --\\
\midrule
\multicolumn{7}{@{}l}{\textit{8-bit}}\\[1pt]
AEIC, ft8 & base codec & 0.017 & 22.04 & 16.92 & $-5.13$ & --\\
 & ours, full method & 0.017 & 22.04 & 22.04 & 0.00 & --\\
 & ours, closest candidate & 0.017 & 22.04 & 20.95 & $-1.09$ & 1,200\\
 & ours, unrepaired output & 0.017 & 22.04 & 16.60 & $-5.44$ & 1,200\\
HPCM, $\lambda{=}.013$ & base codec & 0.378 & 33.99 & 28.12 & $-5.87$ & --\\
 & ours, full method & 0.378 & 33.95 & 33.95 & 0.00 & --\\
 & ours, closest candidate & 0.378 & 33.95 & 32.51 & $-1.44$ & 1,200\\
 & ours, unrepaired output & 0.378 & 33.96 & 30.33 & $-3.63$ & 1,200\\
MSINN, anchor 2 & trained & 0.294 & 32.10 & 30.00 & $-2.10$ & --\\
MSINN, anchor 3 & trained & 0.429 & 33.88 & 31.90 & $-1.97$ & --\\
Li (exact), q3 & trained & 0.399 & 31.66 & 10.75 & $-20.91$ & --\\
\bottomrule
\end{tabular}
\end{table*}

\begin{table*}[t]
\caption{\small\textbf{NLBP configurations of the reported results.} Every sampling step applies the correction; $\lambda_t$ takes its early value, then its late value. $\omega$ is the start-noise weight of the decoder-trade-off figure ($s_0$ in the teaser figure). $K$: Adam steps of the final $b$ refinement (learning rate .0021 for ELIC rates 1--4, otherwise .003). Seeds derive from the received symbols; $D_{\mathrm{ref}}$ is the codec's own synthesis. 8-bit rows apply the 8-bit repair and the stored-image code check; float rows check the code on the float output and are scored on an exported 8-bit image. FFHQ tuning used 100 images disjoint from the test set and the frozen rule ``lowest FID$_{64}$ among settings with PSNR at least Xu's minus 0.3\,dB''; Kodak settings were chosen on the odd-numbered images.}\par\vspace{-0.1cm}
\label{tab:nlbp-configs}
\centering\scriptsize
\setlength{\tabcolsep}{2.5pt}
\renewcommand{\arraystretch}{1.08}
\begin{tabular*}{\textwidth}{@{\extracolsep{\fill}}>{\raggedright\arraybackslash}p{0.17\textwidth}llrrlrlrl>{\raggedright\arraybackslash}p{0.085\textwidth}@{}}
\toprule
Reported result & Codec, rate & Prior & Steps & $\omega$ & $\lambda_t$ & $\kappa$ & $v_{\mathrm{ref}}$ & $K$ & Storage & Selection\\
\midrule
\multicolumn{11}{@{}l}{\textit{ELIC on FFHQ, prior \texttt{ffhq\_10m}}}\\[1pt]
FFHQ first decoding; teaser endpoint & rate 1 (.080) & ffhq\_10m & 318 & .57 & .114 & .07 & $\hat y$ & 30 & 8-bit & tuned (final)\\
FFHQ first decoding; teaser endpoint & rate 2 (.115) & ffhq\_10m & 251 & .49 & .114 & .10 & $\hat y$ & 10 & 8-bit & tuned (final)\\
FFHQ first decoding; teaser endpoint & rate 3 (.164) & ffhq\_10m & 318 & .57 & .114 & .15 & $\hat y$ & 30 & 8-bit & tuned (final)\\
FFHQ first decoding; 50 rounds; teaser endpoint & rate 4 (.372) & ffhq\_10m & 251 & .49 & .114 & .36 & $\hat y$ & 10 & 8-bit & tuned (final)\\
FFHQ first decoding; teaser endpoint & rate 5 (.661) & ffhq\_10m & 201 & .67 & .50$\to$.17 & .85 & $\hat y$ & 10 & 8-bit & tuned (final)\\
Trade-off endpoints (b) & rates 1--4 & ffhq\_10m & 951 & .99 & .286$\to$.114 & as above & $\hat y$ & 30 & float & tuned (earlier)\\
Trade-off endpoint (b); 100 rounds & rate 5 (.661) & ffhq\_10m & 276 & .82 & .50$\to$.17 & .85 & $\hat y$ & 10 & float; 8-bit & tuned (earlier)\\
Intermediate points (teaser, trade-offs b) & rates 1--5 & ffhq\_10m & 51 (26) & .15 & .114 (.17) & as above & $\hat y$ & as above & float & untuned\\
\midrule
\multicolumn{11}{@{}l}{\textit{HPCM on Kodak}}\\[1pt]
Generator choice; controls; trade-off endpoint (a) & $\lambda$ .0018 & PixNerd-XXL-P16 & 45 & .45 & 1.0$\to$.345 & .4 & $y^\star$ & 10 & float & start on odd images\\
Controls (other starts) & $\lambda$ .0018 & PixNerd-XXL-P16 & 15--77 & .20--.65 & 1.0$\to$.345 & .4 & $y^\star$ & 10 & float & start on odd images\\
Teaser; trade-offs (a), other rates & $\lambda$ .0035--.013 & PixNerd-XXL-P16 & 11--45 & .15--.45 & 1.0$\to$.345 & .4 & $y^\star$ & 10 & float & untuned\\
Controls & $\lambda$ .0018 & Z-Image & 23 & .45 & 1.0$\to$.345 & 1.0 & $\hat y$ & 10 & float & start on odd images\\
Generator choice & $\lambda$ .0018 & Z-Image & 23 & .45 & 1.0$\to$.345 & 1.0 & $\hat y$ & 10 & float & untuned\\
Generator choice & $\lambda$ .0018 & FLUX.2 [dev]$^{a}$ & 23 & .45 & 1.0$\to$.345 & 1.0 & $\hat y$ & 10 & float & untuned\\
\midrule
\multicolumn{11}{@{}l}{\textit{AEIC on Kodak, rate ft2}}\\[1pt]
Controls & ft2 & Z-Image$^{a}$ & 1 & .01 & .30 & .64 & $y^\star$ & 40 & float & odd images\\
Controls & ft2 & DeCo-512 & 1 & .15 & .003 & .64 & $y^\star$ & 30 & float & odd images\\
Controls & ft2 & DeCo-512 & 4 & .61 & .45--.60 & .64 & $\hat y$ & 50--200 & float & odd images\\
Controls & ft2 & DeCo-512 & 4 & .64 & .41--.42 & .64 & $\hat y$ & 50--200 & float & odd images\\
\bottomrule
\multicolumn{11}{@{}l}{$^{a}$Prompt ``a high-quality photograph''; FLUX.2 [dev] uses embedded guidance 4.0. Other priors use an empty prompt or the null class.}\\
\end{tabular*}
\end{table*}

\begin{table*}[t]
\caption{\small\textbf{First-pass cost as BD-rate against the base codec.} Round 1; adapted and base decoding of the same first file at all tested rates. BD-rate: average rate change at equal PSNR (positive: more bits); BD-PSNR: average PSNR change at equal rate; both from cubic fits of the unrounded image means. TCM uses its rates $\lambda\le.025$; GLC rates are estimated. Settings with fewer than four rates are omitted.}
\label{tab:bd-rate}
\centering\small
\begin{tabular}{@{}llrrr@{}}
\toprule
Protocol, data & Codec & Rates & BD-rate (\%) & BD-PSNR (dB)\\
\midrule
Float, Kodak & AEIC & 5 & $+$0.04 & $-$0.001\\
 & HPCM & 6 & $+$1.13 & $-$0.047\\
 & ELIC & 6 & $+$0.14 & $-$0.005\\
 & GLC & 4 & $-$0.33 & $+$0.007\\
Float, DIV2K & AEIC & 5 & $+$0.51 & $-$0.010\\
Clipped float, Kodak & AEIC & 5 & 0.00 & 0.000\\
 & HPCM & 4 & $+$0.84 & $-$0.031\\
8-bit, Kodak & AEIC & 5 & 0.00 & 0.000\\
 & HPCM & 6 & $+$1.60 & $-$0.069\\
 & ELIC & 6 & $+$0.28 & $-$0.011\\
 & MS-ILLM & 5 & $+$0.17 & $-$0.004\\
 & TCM & 5 & $+$0.34 & $-$0.014\\
 & Hyperprior & 7 & $+$5.36 & $-$0.276\\
 & Mean-scale & 7 & $+$4.32 & $-$0.210\\
8-bit, CLIC (41) & AEIC & 5 & 0.00 & 0.000\\
8-bit, DIV2K (100) & AEIC & 5 & 0.00 & 0.000\\
\bottomrule
\end{tabular}
\end{table*}

\fi
\end{document}